\documentclass[journal]{IEEEtran}
\usepackage[T1]{fontenc} 

\usepackage{amsthm}
\usepackage{amsmath}
\usepackage{amssymb}

\usepackage{graphicx}
\graphicspath{{figures/}}
\usepackage{epstopdf}
\usepackage{caption}
\usepackage{subfigure}
\usepackage{multirow}
\usepackage{booktabs}
\usepackage{adjustbox}
\usepackage{diagbox}
\usepackage{tabularx}
\usepackage{array}
\usepackage[table]{xcolor}

\usepackage{cite}

\usepackage{algorithmicx}
\usepackage[ruled,linesnumbered]{algorithm2e}

\usepackage{url}
\usepackage{siunitx}
\usepackage{exscale}
\usepackage{relsize}
\usepackage{fixltx2e} 
\usepackage{ltxcmds}  
\usepackage{subscript}
\usepackage{hyperref}
\hypersetup{hidelinks}

\usepackage{bm}
\usepackage{nomencl}
\makenomenclature
\usepackage{threeparttable}

\makeatletter
\newcommand*\bigcdot{\mathpalette\bigcdot@{.5}}
\newcommand*\bigcdot@[2]{\mathbin{\vcenter{\hbox{\scalebox{#2}{$\m@th#1\bullet$}}}}}
\makeatother

\theoremstyle{definition}

\newtheoremstyle{myTheo_style1}%
  {3pt}%
  {0pt}%
  {}%
  {}%
  {\itshape}%
  {}%
  {5pt}%
  {}%

\newtheoremstyle{myTheo_style2}%
  {3pt}%
  {0pt}%
  {}%
  {}%
  {\itshape}%
  {:}%
  {5pt}%
  {}%

\newtheoremstyle{myProof_style}%
  {0pt}%
  {0pt}%
  {}%
  {}%
  {\itshape}%
  {:}%
  {5pt}%
  {}%

\newtheoremstyle{myAssumption_style}%
  {0pt}%
  {0pt}%
  {}%
  {}%
  {\itshape}%
  {:}%
  {5pt}%
  {}%
\newtheorem{myTheo}{Theorem} 

\theoremstyle{myTheo_style1}
\newtheorem{myTheo1}[myTheo]{Theorem}

\theoremstyle{myTheo_style2}

\theoremstyle{myTheo_style1}
\theoremstyle{myProof_style}
\newtheorem{myCorollary}{Corollary}
\theoremstyle{myAssumption_style}

\theoremstyle{myProof_style}
\newtheorem*{myProof}{Proof}
\theoremstyle{myProof_style}

\ifCLASSINFOpdf

\else

\fi

\begin{document}

\title{PB-OEL: A Performance-Bounded Online Ensemble Learning Framework With Mixed Feedback for \\Real-Time Safety Assessment}

\author{Songqiao Hu, 
        Zeyi Liu, 
        Lufeng Hao, Yinzhong Cheng, 
        and~Xiao He,~\IEEEmembership{Senior~Member,~IEEE}
\thanks{This work was supported in part by the National Natural Science Foundation of China under Grants 62525308, 624B2087, 62473223, and 52172323, and in part by the Beijing Natural Science Foundation under Grant L241016 (Corresponding authors: Xiao He, Yinzhong Cheng).}
\thanks{Songqiao Hu, Zeyi Liu, and Xiao He are with the Department of Automation, and the Institute for Embodied Intelligence and Robotics, Tsinghua University, Beijing 100084, China (e-mails: hsq23@mails.tsinghua.edu.cn; liuzy21@mails.tsinghua.edu.cn; hexiao@tsinghua.edu.cn).

Lufeng Hao and Yinzhong Cheng are with China Ship Research and Development Academy, Beijing 100192, China            (e-mails: haolufeng@csrda.com; chengyinzhong@csrda.com).
}}


\maketitle
\begin{abstract}
Real-time safety assessment is critical for ensuring the reliable operation of complex dynamic systems. However, obtaining full safety labels in real-time is often prohibitively expensive, resulting in a challenging mixed-feedback scenario dominated by partial feedback, especially under concept drift. Furthermore, existing online ensemble methods typically rely on heuristic weight allocation, lacking provable performance guarantees under such limited-feedback conditions. To address these challenges, we propose PB-OEL, a performance-bounded online ensemble learning framework designed for real-time safety assessment under mixed feedback. At the ensemble level, a theoretical framework is established to bound the performance of the ensemble classifier relative to its base classifiers across varying feedback ratios. By formally defining the form of expert advice, the bound guarantees that the ensemble outperforms any individual base classifier over a sufficiently large data stream. At the base-classifier level, a penalty-based update strategy is introduced, enabling base models to explicitly leverage misclassified samples rather than simply discarding them. Extensive experiments on the real-world \textit{Jiaolong} manned submersible dataset demonstrate that PB-OEL maintains robust predictive performance and outperforms state-of-the-art methods.
\end{abstract}




\section{Introduction}

\IEEEPARstart{R}{eal-time} safety assessment (RTSA) is essential for dynamic systems, where safety is defined as the capability of the system to operate without causing harm to personnel, equipment, or the environment \cite{he2024real}. In real-world practice, such systems are increasingly monitored through multi-source sensor streams \cite{yan2024multisensor}.
The necessity of RTSA becomes evident when these systems operate in non-stationary environments, as exemplified by deep-sea manned submersibles \cite{liu2022online}. During a mission, onboard systems continuously generate multi-source sensor data. RTSA utilizes these streaming data to assess the safety level of the submersible, thereby providing  timely alerts and decision support.

However, a fundamental bottleneck in deploying RTSA is the prohibitive cost of acquiring complete annotations to continuously update the assessment model \cite{qian2024deep,liu2022online,he2023dynamic}. During the online assessment phase, requiring domain experts to determine the exact safety level for every sample imposes an overwhelming burden. While active learning alleviates this by selecting only the most informative samples for annotation, it still demands a full label for every queried instance \cite{cacciarelli2024active,yan2024incremental}. Unfortunately, in complex or highly ambiguous situations, identifying the exact safety level is often exceptionally challenging. Since providing full labels is realistic only for a small fraction of simple or obvious samples, standard active learning may not be well-suited for such scenarios. Instead, simply verifying the correctness of a given prediction significantly reduces the cognitive load on experts \cite{zhou2018brief}. Driven by this reality, this paper focuses on a practical interaction mode: the assessment model proactively outputs a predicted safety level, and primarily receives binary feedback regarding its correctness, supplemented by full labels for only a minimal fraction of unambiguous samples. Under this realistic mixed-feedback setting, we explore how to continuously and robustly adapt the assessment model, thereby enhancing predictive performance with minimal human effort.

While the mixed-feedback setting significantly reduces cognitive load, it introduces a critical challenge for the assessment model when deployed in non-stationary environments. Specifically, when \textit{concept drift} occurs, the model becomes prone to misclassifications based on outdated patterns \cite{lu2018learning, lin2025uncertainty,hu2024cadm,cao2023online}. Under partial feedback, these misclassifications yield only  negative confirmations without revealing the true labels necessary for model updates \cite{ishida2017learning}. Consequently, the model fails to adapt to new concepts, triggering a vicious cycle of persistent errors. Therefore, it is imperative to develop a robust learning framework capable of synergistically utilizing rare full labels, positive confirmations, and negative complementary feedback to ensure continuous adaptation in data streams.

Online ensemble learning is a powerful paradigm for non-stationary data streams because it combines multiple base classifiers and dynamically adjusts their contributions \cite{krawczyk2017ensemble,liu2023robust,tanveer2023ensemble,campagner2024ensemble}. Existing online ensemble methods typically update classifier weights based on predictive performance, error rates, classifier age, or drift detector outputs \cite{zhang2019online,lu2019adaptive,ren2018knowledge,cano2020kappa,jiao2022dynamic}. However, most of these approaches assume the availability of true labels after prediction. Under the aforementioned mixed-feedback setting, particularly when predictions are rejected without revealing the correct class, critical mechanisms such as drift adaptation, base-classifier updates, and weight allocation are severely hindered. Moreover, many weight allocation strategies remain heuristic and lack rigorous performance guarantees under such scenarios \cite{zhang2019online,ren2018knowledge,cano2020kappa,wilsonmulti,jiao2022dynamic}. Even when theoretical bounds are provided, they often lack clear physical interpretability or remain computationally intractable \cite{tekin2016adaptive,pang2018dynamic}.

To address these challenges, we propose a mixed-feedback \textbf{P}erformance-\textbf{B}ounded \textbf{O}nline \textbf{E}nsemble \textbf{L}earning (PB-OEL) framework for real-time safety assessment. At the ensemble level, by treating base classifiers as experts, PB-OEL integrates Hedge-style full-information updates with EXP4-inspired bandit updates to dynamically adjust classifier weights \cite{freund1997decision,auer2002nonstochastic}. At the base-classifier level, we incorporate a penalty-based update strategy, enabling base models to explicitly leverage negative partial feedback rather than discarding it. Crucially, we derive a theoretical lower bound on the expected accuracy of the ensemble classifier. The main contributions of this paper are summarized as follows:
\begin{enumerate}
    \item A unified mixed-feedback online ensemble framework is proposed for real-time safety assessment. The framework seamlessly integrates Hedge-style full-information updates with EXP4-inspired bandit updates to handle non-stationary data streams under varying full-to-partial feedback ratios.
    
    \item A theoretical lower bound on the expected accuracy on the ensemble classifier is established. It is mathematically proven that, across varying feedback ratios, the expected accuracy of the ensemble exceeds that of any individual base classifier over a sufficiently large data stream.
    
    \item A penalty-based update strategy is introduced at the base-classifier level. Instead of discarding misclassified samples, this mechanism enables base models to explicitly leverage negative partial feedback for performance improvement.
\end{enumerate}

The remainder of this paper is organized as follows. Section \ref{sec:problem} formulates the problem, while Section \ref{sec:methods} details the proposed methodology. Section \ref{sec:application} reports the experimental results. Finally, Section \ref{sec:conclusion} concludes the work.

\begin{figure*}[htbp]
\centering
    \includegraphics[width=0.94\textwidth]{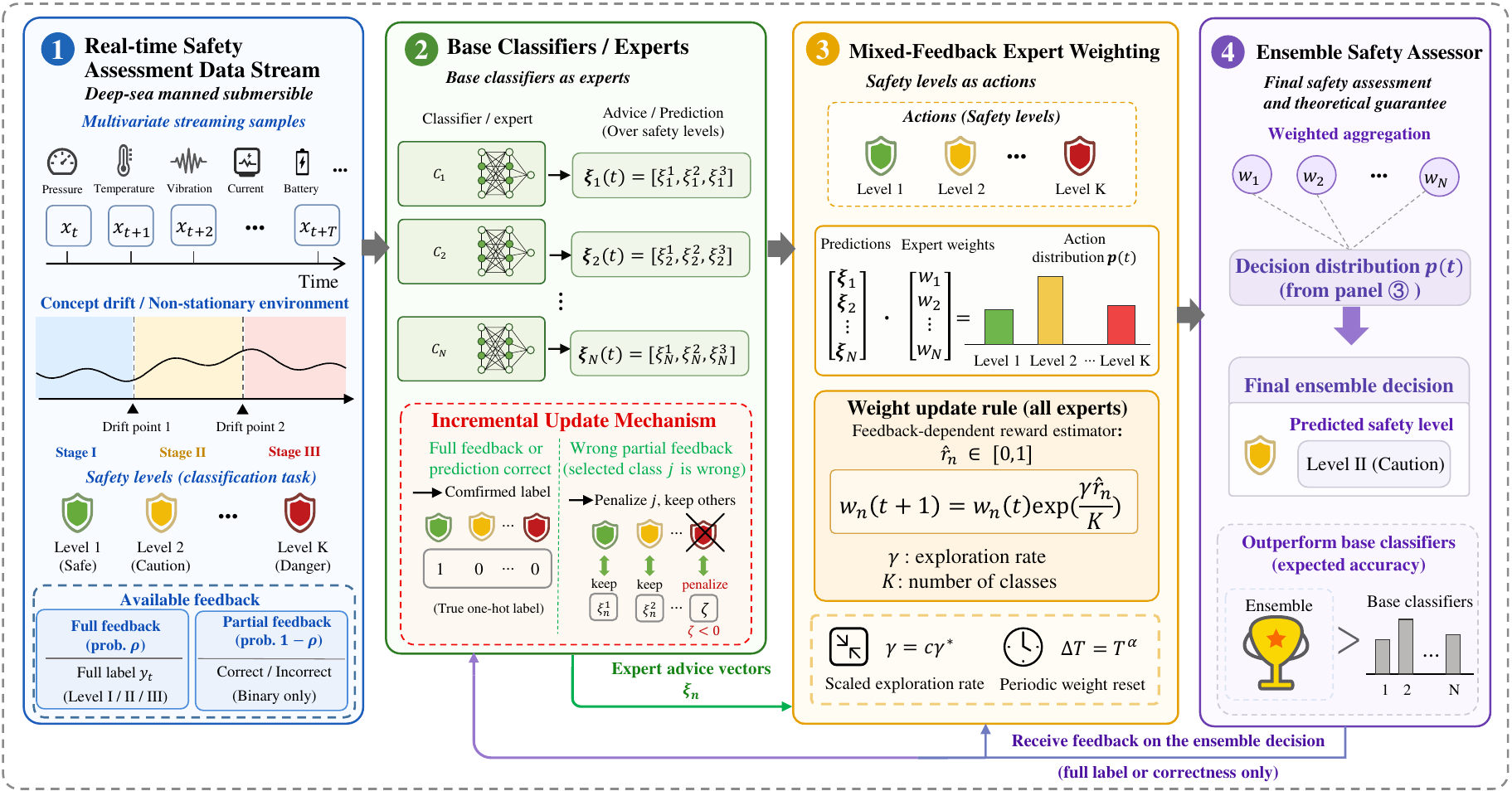}
        \caption{Overall architecture of the proposed PB-OEL framework. (1) The system processes non-stationary streaming data under mixed-feedback conditions. (2) Base classifiers act as experts to provide prediction advice, continuously adapting via a penalty-based incremental update mechanism. (3) Expert weights are dynamically optimized using a mixed-feedback estimator, integrating a scaled exploration rate and a periodic restart strategy. Conceptually, the predicted classes are formulated as actions, which equivalently correspond to the arms of a multi-armed bandit. (4) The ensemble aggregates these weighted decisions to output the final prediction, establishing a theoretical performance guarantee.}
        \label{fig:setting}
\end{figure*}

\section{Problem Formulation}
\label{sec:problem}
\subsection{RTSA Setting under Mixed Feedback}

Let $\mathcal{D}$ denote a data stream generated by a dynamic system, where each instance $\bm{x}_t \in \mathbb{R}^{d}$ at time $t$ comprises $d$ monitored safety indicators. In the offline stage, a small set of labeled samples is available to initialize the assessment model. Let $y_t \in \{1, 2, \ldots, M\}$ represent the true safety level of $\bm{x}_t$, where $M$ is the total number of safety levels. During the online stage, due to limited annotation budgets, only a small fraction of samples receive full feedback, revealing their true labels $y_t$. Let $\rho$ denote the probability of a sample receiving full feedback. For the remaining samples, only partial feedback is provided, indicating whether the current prediction is correct.

To formally characterize this mixed-feedback mechanism, we introduce a random variable $Z_t \in \{\text{F}, \text{P}\}$ to indicate the feedback mode at each time step $t$. Specifically, the full feedback mode denoted by $Z_t = \text{F}$ occurs with probability $\rho$, and the partial feedback mode denoted by $Z_t = \text{P}$ with probability $1-\rho$. The assessment model $\Phi_t$ is incrementally updated based on the limited feedback information. The objective is to maximize the cumulative number of correct predictions under the mixed-feedback constraint.

\subsection{Online Ensemble Learning with Expert Advice}

To tackle the mixed-feedback RTSA task with performance guarantees, we cast the OEL paradigm as an expert-advice multi-armed bandit (MAB) problem \cite{auer2002nonstochastic}. Let $\mathcal{T} = \{1, 2, \ldots , T\}$ denote the time horizon. We treat the $N$ base classifiers as experts $\mathcal{N} = \{1, \ldots , N\}$, and the $M$ safety levels as arms $\mathcal{K} = \{1, \ldots , K\}$, where $K = M$. 

At each time step $t$, given sample $\bm{x}_t$, each expert $n$ outputs an \emph{advice vector} $\bm{\xi}_n(t) \in [0, 1]^K$ indicating its classification confidence. The ensemble  then selects an \emph{action} $a_t \in \mathcal{K}$ as the final prediction. We define the true reward vector $\bm{\mu}_t \in \{0, 1\}^K$, whose $k$-th component is:
\begin{equation}
\label{eq:reward_latent}
\mu_t(k) = \mathbf{1}\{k = y_t\}.
\end{equation}

The ensemble receives an immediate reward $\mu_t(a_t)$ for this action. Crucially, the observability of $\bm{\mu}_t$ depends on $Z_t$: under full feedback ($Z_t = \text{F}$), $\bm{\mu}_t$ is fully revealed, whereas under partial feedback ($Z_t = \text{P}$), only $\mu_t(a_t)$ is observed.

Consequently, the expected reward for expert $n$ is formally defined as:
\begin{equation}
\label{eq:mu_n_t}
r_n(t)=\sum\limits_{k=1}^K \xi_{n}^k(t) \mu_t(k).
\end{equation}

Since $\bm{\mu}_t$ is only partially observed when $Z_t = \text{P}$, exact computation of $r_n(t)$ is infeasible, necessitating robust reward estimators. To overcome this partial observability, our PB-OEL framework unifies the full-information \emph{Hedge} and bandit-feedback \emph{EXP4} algorithms based on the feedback ratio $\rho$. This framework integrates theoretically guaranteed ensemble weight allocation with adaptive base-classifier updates under limited feedback.

\section{Performance-Bounded Online\\ Ensemble Learning Framework}
\label{sec:methods}
In this section, we detail the proposed PB-OEL framework. As illustrated in Fig. \ref{fig:setting}, it operates at two levels: mixed-feedback weight allocation for the ensemble, and penalty-based incremental updates for base classifiers.

\subsection{Weight Allocation Strategy for Base Classifiers}

Optimal weight allocation is paramount for maximizing the generalization capability of the ensemble classifier. As the core of PB-OEL, we extend the classical EXP4 algorithm \cite{lattimore2020bandit} to dynamically accommodate varying feedback modes.

At $t=1$, all base classifiers are initialized with equal weights $w_n(1)=1$. Upon receiving a new sample $\bm{x}_t$, PB-OEL computes the normalized weight for each expert $n$ as $q_n(t) = w_n(t) / \sum_{j=1}^N w_j(t)$. The aggregated advice for class $k$ is $s_k(t) = \sum_{n=1}^N q_n(t) \xi_{n}^k(t)$. 

The determination of action $a_t$ inherently depends on the feedback mode $Z_t$, which is set to be revealed upon the arrival of the sample. Under full feedback $Z_t = \text{F}$, PB-OEL relies entirely on exploitation, setting the sampling distribution to $\bm{p}(t) = \bm{s}(t)$. Conversely, under partial feedback $Z_t = \text{P}$, an exploration term modifies this distribution to:
\begin{equation}
\label{eq:p_k}
p_k(t) = (1-\gamma) s_k(t) + \frac{\gamma}{K},
\end{equation}
where $\gamma \in (0, 1]$ is the exploration rate. Finally, the ensemble draws the action as $a_t \sim \bm{p}(t)$.

Upon receiving feedback, PB-OEL constructs a reward estimator $\hat{r}_n(t)$ for each expert $n$, conditioned on $Z_t$:
\begin{equation}
\label{eq:mu_hat_n}
\hat{r}_n(t) = 
\begin{cases} 
\xi_{n}^{y_t}(t), & \text{if } Z_t = \text{F}, \\ 
\sum\limits_{k=1}^K \xi_{n}^k(t) \frac{\mathbf{1}\{a_t = k\} \mu_t(a_t)}{p_k(t)}, & \text{if } Z_t = \text{P}. 
\end{cases}
\end{equation}

Under full feedback, the exact reward $r_n(t)$ is directly recovered regardless of $a_t$. Under partial feedback, importance sampling ensures unbiasedness, i.e., $\mathbb{E}_{a_t}[\hat{r}_n(t) \mid Z_t=\text{P}] = r_n(t)$. Finally, the expert weights are updated exponentially:
\begin{equation}
\label{eq:w_n_update}
w_n(t+1) = w_n(t) \exp \left(\frac{\gamma \hat{r}_n(t)}{K}\right).
\end{equation}

By unifying these mixed-feedback updates, we first establish a theoretical performance bound for the standard policy (i.e., the precursor to the proposed PB-OEL framework), as formally stated in Theorem \ref{Theorem:ACC}.

\begin{myTheo1}
\label{Theorem:ACC}
Suppose the standard policy operates with exploration rate $\gamma=\min\left\{1, \sqrt{\frac{K \ln N}{T {B_\rho}}}\right\}$, where $B_\rho = \frac{\rho}{8K} + (1-\rho)(e-1)$. Then, the expected accuracy (ACC) of the ensemble satisfies the lower bound:
\begin{equation}
\label{eq:ACC}
\mathbb{E}\left[\text{ACC}_{\text{Std}}\right] \geq {\max_{1\le n \le N}\frac{1}{T}\sum_{t=1}^T \xi_{n}^{y_t}(t)} - {2\sqrt{\frac{B_\rho K \ln N}{T}}},
\end{equation}
where $\xi_{n}^{y_t}(t)$ denotes the true-class confidence of classifier $n$. If the advice vector uses hard voting (i.e., one-hot format where the predicted class is assigned $1$ and others $0$), the ensemble's expected accuracy asymptotically matches or exceeds the accuracy $\text{ACC}_{n^\star}$ of the best base classifier in hindsight:
\begin{equation}
\label{eq:ACC_limit}
\lim_{T \to \infty} \left( \mathbb{E}\left[\text{ACC}_{\text{Std}}\right] - \text{ACC}_{n^\star} \right) \geq 0.
\end{equation}
\end{myTheo1}
\begin{myProof}
See Appendix~\ref{appendix:1}.
\hfill$\blacksquare$
\end{myProof}

Theorem \ref{Theorem:ACC} establishes the performance bound for the standard policy over a fixed horizon using the theoretically optimal exploration rate $\gamma^\star=\sqrt{K\ln N/(TB_\rho)}$.  However, practical deployments in continuous data streams face two primary challenges. First, frequent concept drifts render accumulated historical weights obsolete, severely degrading ensemble performance. Second, although $\gamma^\star$ strictly minimizes the worst-case regret bound, enforcing such extensive exploration in classification tasks often introduces excessive prediction variance and degrades empirical performance.

To simultaneously maintain agile adaptation and optimize empirical accuracy, we introduce two practical adaptations that jointly constitute the complete PB-OEL framework. Inspired by restarting strategies in non-stationary bandits \cite{besbes2014stochastic}, we incorporate a periodic restart mechanism, yielding a variant termed REXP4. Specifically, this mechanism resets the expert weights to $1$ at intervals of $\Delta T = T^\alpha$ ($\alpha \in (0, 1]$), partitioning the data stream into a sequence of $m_T$ batches. Furthermore, to suppress excessive random sampling, we introduce an empirical scaling factor $c \in (0, 1]$ to reduce the optimal exploration rate, i.e., $\gamma = c \gamma^\star$. These integrated mechanisms establish a scaled piecewise bound, as demonstrated in Corollary \ref{corollary:restart}.

\begin{myCorollary}
\label{corollary:restart}
Suppose PB-OEL operates with restart intervals $\Delta_T = T^\alpha$ and a scaled exploration rate $\gamma = c \sqrt{K \ln N / (\Delta T B_\rho)}$. The expected accuracy of the ensemble satisfies the following lower bound:
\begin{equation}
\label{eq:ACC_restart}
\begin{aligned}
\mathbb{E}\left[\text{ACC}_{\text{PB-OEL}}\right] &\geq \underbrace{\frac{1}{T}\sum_{j=1}^{m_T} \max_{1\le n \le N} \sum_{t \in \mathcal{T}_j} \xi_{n}^{y_t}(t)}_{\text{{Piecewise\:Ultimate\:Bound}}} \\
&\quad - \underbrace{(c + c^{-1})\sqrt{B_\rho K \ln N} \left(T^{\frac{\alpha}{2}-1} + T^{-\frac{\alpha}{2}}\right)}_{\text{{Scaled\:Piecewise\:Regret}}}.
\end{aligned}
\end{equation}
\begin{myProof}
    See Appendix~\ref{appendix:2}.
    \hfill$\blacksquare$
\end{myProof}

\end{myCorollary}

This restart strategy reveals a fundamental trade-off: it achieves a higher, more flexible {\itshape Piecewise Ultimate Bound} (as the optimal expert $\max_{n}$ can dynamically switch between batches to adapt to concept drifts), but incurs a greater regret penalty in regret compared to the bound of the standard scaled policy. In long-horizon data streams, where the regret asymptotically vanishes, achieving a higher \textit{Ultimate Bound} is more advantageous from an overall perspective.

\begin{myCorollary}\label{coro:comparison}By incorporating the empirical scaling factor $c$ into Theorem \ref{Theorem:ACC}, the expected accuracy lower bound for the scaled standard policy over the entire horizon $T$ is given by:
\begin{equation}
\label{eq:ACC_scaled_std}
\begin{aligned}
\mathbb{E}[\text{ACC}_{\text{Scaled-Std}}] &\geq \underbrace{\frac{\max\limits_{1 \leq n \leq N}\sum\limits_{t=1}^T \xi_{n}^{y_t}(t)}{T}}_{\text{{{Standard}}{\:Ultimate\:Bound}}} \\
&\quad - \underbrace{\frac{(c+c^{-1}) \sqrt{B_\rho K \ln N}}{\sqrt{T}}}_{\text{{{Scaled\:Standard}}{\:Regret}}}.
\end{aligned}
\end{equation}

Compared to this scaled standard bound, PB-OEL achieves a higher {\itshape Ultimate Bound}, but at the cost of increased {\itshape Regret}.
\end{myCorollary}
\begin{myProof}
See Appendix~\ref{appendix:3}.$\hfill\blacksquare$
\end{myProof}

\subsection{Update Strategy for Base Classifiers}

To enable rapid, real-time incremental updates without storing historical data in streaming scenarios, we adopt the Random Vector Functional-Link (RVFL) network as the base classifier \cite{pao1994learning}.

Given a small initial dataset of state samples $\{\bm{x}_i\}_{i=1}^N$ with $\bm{x}_i \in \mathbb{R}^{1 \times n}$ and their one-hot target labels $\tilde{\bm{y}}_i \in \mathbb{R}^{1 \times K}$, let $\tilde{\bm{x}}_i \in \mathbb{R}^{1 \times (n+L)}$ denote the extended feature vector concatenating the original state $\bm{x}_i$ and its enhancement features. Generated via $N_1$ groups of $N_2$ nodes ($L = N_1 N_2$), the extended vector is compactly constructed as:
\begin{equation}
\tilde{\bm{x}}_i = \left[ \bm{x}_i, \; \phi(\bm{x}_i \bm{W}_{e_1} + \bm{b}_{e_1}), \cdots, \phi(\bm{x}_i \bm{W}_{e_{N_1}} + \bm{b}_{e_{N_1}}) \right],
\end{equation}
where $\phi(\cdot)$ is the element-wise activation function. For each group $j \in \{1, \dots, N_1\}$, the weight matrix $\bm{W}_{e_j} \in \mathbb{R}^{n \times N_2}$ and bias $\bm{b}_{e_j} \in \mathbb{R}^{1 \times N_2}$ are randomly initialized and kept fixed.

Define the extended data matrix $\bm{A} \in \mathbb{R}^{N \times (n+L)}$ and the label matrix $\bm{Y} \in \mathbb{R}^{N \times K}$ as:
\begin{equation}
\bm{A} = [\tilde{\bm{x}}_1^\top, \tilde{\bm{x}}_2^\top, \cdots, \tilde{\bm{x}}_N^\top]^\top, \quad \bm{Y} = [\tilde{\bm{y}}_1^\top, \tilde{\bm{y}}_2^\top, \cdots, \tilde{\bm{y}}_N^\top]^\top.
\end{equation}

The initial output weight matrix $\bm{W}_b \in \mathbb{R}^{(n+L) \times K}$ is obtained by solving the regularized least-squares problem:
\begin{equation}
\bm{W}_b = \arg\min_{\bm{W}} \lambda \|\bm{W}\|_2^2 + \|\bm{A}\bm{W} - \bm{Y}\|_2^2,
\end{equation}
yielding the closed-form solution:
\begin{equation}
\bm{W}_b = (\lambda \bm{I} + \bm{A}^\top \bm{A})^{-1} \bm{A}^\top \bm{Y}.
\end{equation}

During the online phase, the raw prediction vector for a given sequential sample $\bm{x}_t$ is computed as $\hat{\bm{y}}_t = \tilde{\bm{x}}_t \bm{W}_b \in \mathbb{R}^{1 \times K}$. A normalization step is then applied to $\hat{\bm{y}}_t$ to yield the final prediction confidence vector $\bm{\xi}_t \in (0, 1)^{1 \times K}$.
\newtheorem{lemma}{Lemma}
\begin{lemma}[\hspace{-0.03em}\cite{hager1989updating,liu2023dynamic}]
\label{lemma:woodbury}
Let $\mathbf{P}_t = (\lambda \bm{I} + \bm{A}_t^\top \bm{A}_t)^{-1}$ be the inverse covariance matrix at time step $t$. For a new sequential sample $\tilde{\bm{x}}_t$ and its constructed target label vector $\bm{y}_t^* \in \mathbb{R}^{1 \times K}$, the output weight matrix $\bm{W}_{b, t}$ and the inverse covariance matrix $\mathbf{P}_t$ are incrementally updated without retraining via:
\begin{equation}
\label{eq:gain_k}
\mathbf{K}_t = \frac{\mathbf{P}_{t-1} \tilde{\bm{x}}_t^\top}{1 + \tilde{\bm{x}}_t \mathbf{P}_{t-1} \tilde{\bm{x}}_t^\top},
\end{equation}
\begin{equation}
\label{eq:update_P}
\mathbf{P}_t = \mathbf{P}_{t-1} - \mathbf{K}_t \tilde{\bm{x}}_t \mathbf{P}_{t-1},
\end{equation}
\begin{equation}
\label{eq:update_beta}
\bm{W}_{b, t} = \bm{W}_{b, t-1} + \mathbf{K}_t \left( \bm{y}_t^* - \tilde{\bm{x}}_t \bm{W}_{b, t-1} \right).
\end{equation}
\end{lemma}

Based on the feedback mode $Z_t$ and prediction correctness $\mu_t(a_t)$, the target vector $\bm{y}_t^* = [y_{t,1}^*, \dots, y_{t,K}^*] \in \mathbb{R}^{1 \times K}$ is conditionally constructed. If $Z_t = \text{F}$ or $\mu_t(a_t) = 1$, the true label $y_t$ is revealed, and $\bm{y}_t^*$ is set to its one-hot encoding. Conversely, if $Z_t = \text{P}$ and $\mu_t(a_t) = 0$, the true label remains unobserved. Simply omitting model updates in this scenario can cause the system to trap in persistent error loops, particularly when concept drift occurs. Fortunately, the feedback explicitly isolates the incorrect action $a_t$. To leverage this critical negative information, we penalize the incorrect class $a_t$ with a predefined negative parameter $\zeta$ while keeping other class confidences unchanged. Formally, the components of $\bm{y}_t^*$ are defined as:
\begin{equation}
y_{t,k}^* = \begin{cases}
\mathbf{1}\{k = y_t\}, & \text{if } Z_t = \text{F} \text{ or } \mu_t(a_t) = 1, \\
\xi_t^k, & \text{if } Z_t = \text{P}, \mu_t(a_t) = 0 \text{ and } k \neq a_t, \\
\zeta, & \text{if } Z_t = \text{P}, \mu_t(a_t) = 0 \text{ and } k = a_t.
\end{cases}
\end{equation}

Applying $\bm{y}_t^*$ to Lemma~\ref{lemma:woodbury} enables real-time adaptation. While sufficient in stationary environments, online streams often suffer from concept drift, rendering incremental updates alone inadequate \cite{sun2024novel,vzliobaite2016overview,zhou2024class}. To detect such shifts, the Hoeffding inequality is applied \cite{hoeffding1994probability,wilson2023homogeneous,li2022high}. Assume that $X_1, \cdots, X_{n_1}, X_{n_1+1}, \cdots, X_{n_1+n_2}$ are independent random variables with values in $[0, 1]$. Let $\overline{X}=\frac{1}{n_1}\sum_{i=1}^{n_1}X_i$ and $\overline{Z}=\frac{1}{n_2}\sum_{i=n_1+1}^{n_1+n_2}X_i$. Then, for any $\varepsilon>0$: 
\begin{equation}
\begin{aligned}
\mathbb{P}\left\{\overline{X}-\overline{Z}-\left(\mathbb{E}[\overline{X}]-\mathbb{E}[\overline{Z}]\right) \geq \varepsilon\right\} \leq e^{-\frac{2 \varepsilon^2 n_1 n_2}{n_1+n_2}}.
\end{aligned}
\end{equation}
 
Generally, the null hypothesis $H_0: \mathbb{E}[\overline{X}] \leq \mathbb{E}[\overline{Z}]$ assumes a stationary environment, whereas $\mathbb{E}[\overline{X}] > \mathbb{E}[\overline{Z}]$ indicates concept drift. Given a confidence level $\delta$, $H_0$ is rejected if 
\begin{equation}
\overline{X}-\overline{Z}\geq\varepsilon_\delta, 
\end{equation}
where $\varepsilon_\delta=\sqrt{n_1\ln ({1}/{\delta})/\left[2 n_1(n_1+n_2)\right]}$.

Consequently, monitoring changes in $\xi_n^{y_t}$ is highly effective because its expectation dictates the theoretical bound of the ensemble. For each base classifier, a sliding window tracks the true-class confidence. The HDDM method \cite{frias2014online}, backed by the Hoeffding inequality, then searches for the optimal cut point to detect drift. If drift is confirmed, the classifier bypasses incremental updates and utilizes the latest samples in the buffer for retraining \cite{liu2022concept}.

\subsection{Overall Procedure of PB-OEL}
Algorithm \ref{alg:pboel} outlines the complete execution flow of PB-OEL. After initializing the base classifiers and uniform ensemble weights, the system sequentially processes each incoming sample $\bm{x}_t$ and its feedback mode $Z_t$. Lines 3-9 detail the prediction process: advice vectors from all base classifiers are aggregated into action probabilities $\bm{p}(t)$, which incorporate an exploration term $\gamma/K$ under partial feedback before sampling the action $a_t$. Upon execution, PB-OEL observes the reward $\mu_t(a_t)$ and obtains $y_t$ (if $Z_t = \text{F}$) to update its weights and construct the conditional target vector $\bm{y}_t^*$. Lines 12-20 detail the base-classifier updates. Specifically, each classifier incrementally updates its output weights using $\bm{x}_t$ and $\bm{y}_t^*$. Crucially, whenever the true label is revealed, it is appended to a local buffer $L_n$ to monitor concept drift via HDDM, triggering retraining if detected. Finally, a periodic reset mechanism reinitializes the ensemble weights every $\Delta_T$ steps to ensure long-term adaptability (Lines 21-24).

\begin{figure}[htbp]
    \centering
    \includegraphics[width=0.85\linewidth]{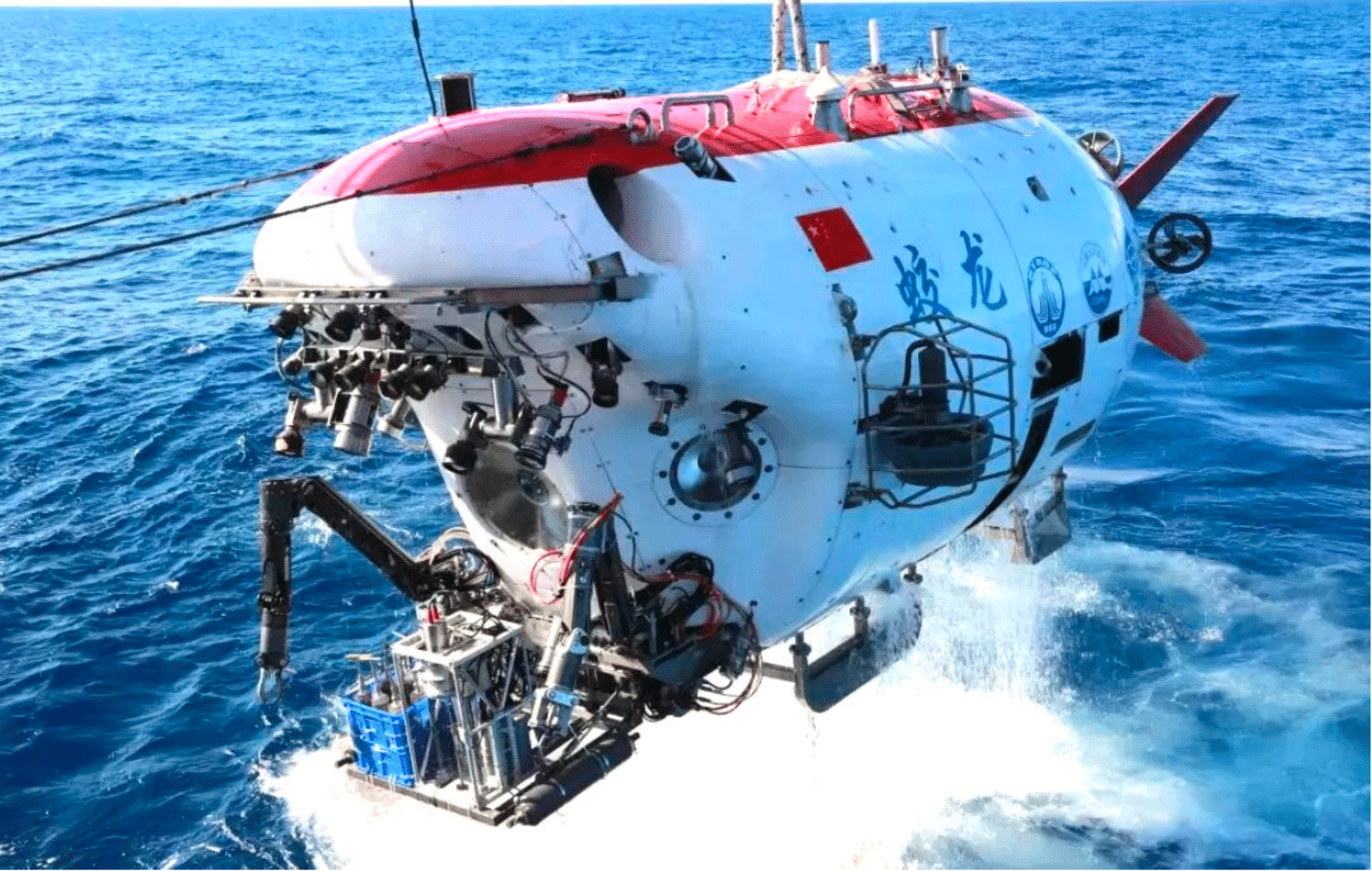}
    \caption{The Jiaolong DSMS in operation during an ocean exploration mission}
    \label{fig:jiaolong}
\end{figure}

\section{Experiments}
\label{sec:application}
\subsection{Background}
To validate our approach, we employ a real-world dataset sourced from the National Deep Sea Center in Qingdao, China. As depicted in Fig.~\ref{fig:jiaolong}, this dataset was collected during the 151st dive mission of the Jiaolong Deep-Sea Manned Submersible (DSMS) on March 19, 2017 \cite{hu2024cadm}. Structured as a multivariate time series, the data is acquired by both the internal life support systems and various external instruments of the submersible, including conductivity, temperature, and depth sensors. Specifically, the monitoring logs record measurements from 24 distinct sensors, tracking critical indicators such as carbon dioxide and oxygen concentrations, attitude angles, thrust, and moments. To reflect the complex underwater operational environment, the collected instances are annotated with three discrete safety levels\footnote{Please refer to the repository for additional information: \url{https://github.com/THUFDD/JiaolongDSMS_datasets}}. Furthermore, as the submersible operates across varying depths, the underlying criteria for determining its safety status shift dynamically, inherently inducing concept drift within the data stream.

\begin{algorithm}[htbp]
\caption{PB-OEL Framework}
\label{alg:pboel}
\SetKwInput{KwData}{Input}
\KwData{Ensemble size $N$, class count $K$, $\Delta_T = T^\alpha$, exploration rate $\gamma$, penalty $\zeta < 0$.}
\SetKwInput{KwInit}{Initialize}
\KwInit{Base classifiers $\{C_n\}_{n=1}^N$ via warm-up, weights $w_n(1) = 1$, buffers $L_n \leftarrow \emptyset$, $\tau = 0$.}

\While{data stream is active at time $t$}{
    Receive sample $\bm{x}_t$ and feedback mode~$Z_t$\\
    Obtain advice vectors $\{\bm{\xi}_{n}(t)\}_{n=1}^N$ from all classifiers\\
    Compute action probabilities $\bm{p}(t)$ via weight aggregation\\
    
    \If{$Z_t = \text{P}$}{
        $\bm{p}(t) \leftarrow (1-\gamma)\bm{p}(t) + \gamma/K$\\
    }
    Sample predicted action $a_t \sim \bm{p}(t)$\\
    
   Execute $a_t$, observe reward $\mu_t(a_t)$, and obtain $y_t$ if $Z_t \!=\! \text{F}$\\
    Estimate rewards $\hat{r}_n(t)$ and update weights $w_n(t+1)$\\
    Construct target vector $\bm{y}_t^*$\\

    \For{$n = 1, \dots, N$}{
        Update  output weights $\bm{W}_b$ of $C_n$ using $\bm{x}_t, \bm{y}_t^*$\\
        \If{$Z_t = \text{F}$ \textnormal{\textbf{or}} $\mu_t(a_t) = 1$}{
            Append true-class confidence $\xi_n^{y_t}(t)$ to $L_n$\\
            \If{\textnormal{HDDM}($L_n$) detects drift}{
               Retrain $C_n$ using recent sample buffer\\
            }
        }
    }
    
    $\tau \leftarrow \tau + 1$\\
    \If{$\tau > \Delta_T$}{
        Reset $\tau \leftarrow 0$ and $w_n(t+1) \leftarrow 1, \forall n$\\
    }
}
\end{algorithm}

\begin{figure*}[h]
    \captionsetup{
    labelfont={color=black},
    textfont={color=black}
}
    \centering
    \begin{minipage}[c]{0.31\linewidth}\footnotesize
        \centering
        \includegraphics[width=\textwidth]{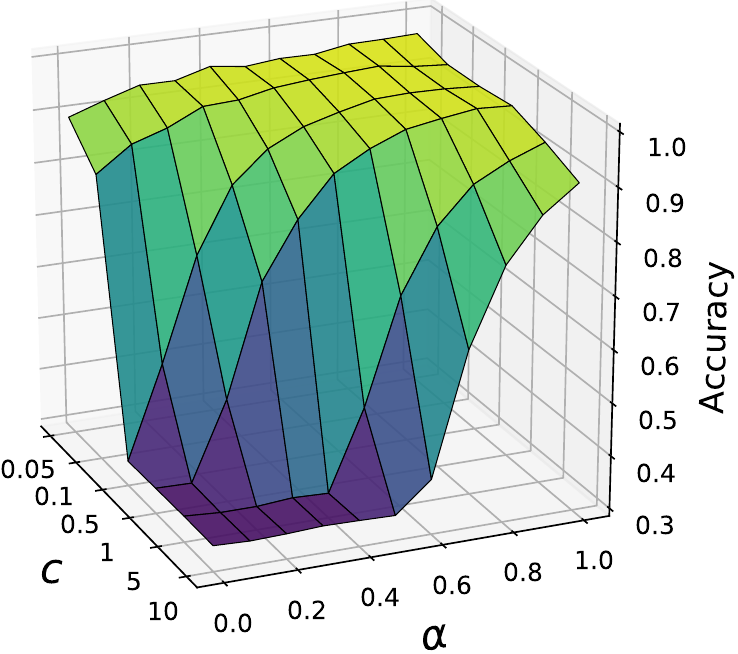}
        \centerline{{(a) $\rho = 0.0$}}
    \end{minipage}\hfill
    \begin{minipage}[c]{0.31\linewidth}\footnotesize
        \centering
        \includegraphics[width=\textwidth]{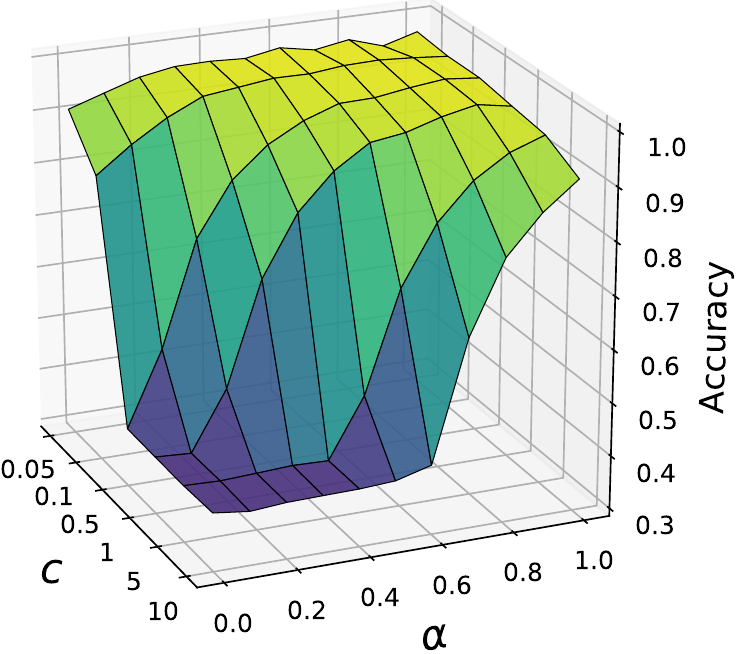}
        \centerline{{(b) $\rho = 0.1$}}
    \end{minipage}\hfill
    \begin{minipage}[c]{0.36\linewidth}\footnotesize
        \centering
        \includegraphics[width=\textwidth]{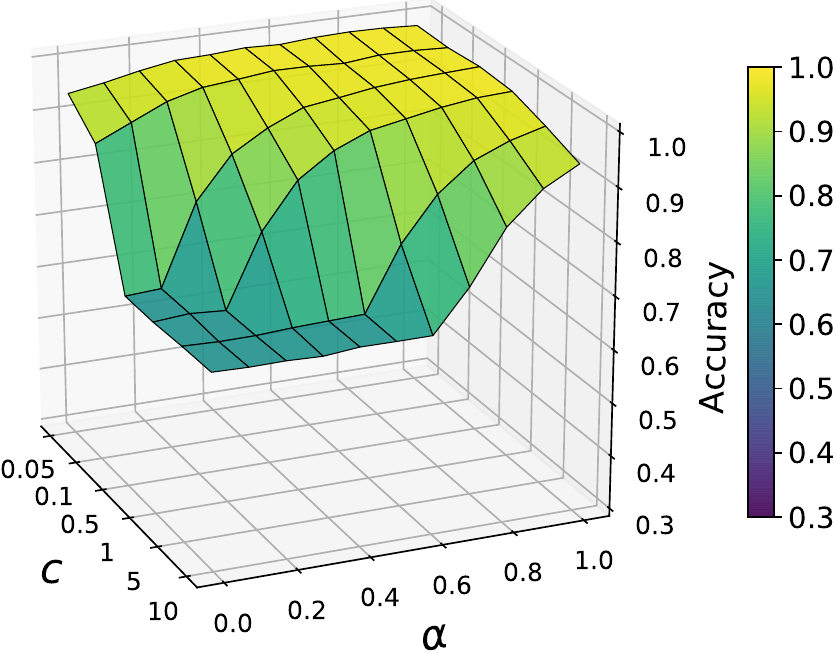}
        \centerline{{(c) $\rho = 0.5$}}
    \end{minipage}\\[6pt]
    \caption{Performance landscapes illustrating the impact of the regret scaling factor (${c}$) and the periodic restart factor (${\alpha}$) on predictive accuracy.}
    \label{fig:parameter}
\end{figure*}
\subsection{Evaluation Metrics}

The cumulative prediction accuracy (Acc) is used as the primary evaluation metric. For a data stream with $T$ instances, it is defined as
\begin{equation}
\text{Acc} = \frac{1}{T} \sum_{t=1}^{T} \mathbb{I}(\hat{y}_t = y_t),
\end{equation}
where $\hat{y}_t$ and $y_t$ denote the predicted and ground-truth labels at time step $t$, respectively.

Each experiment is independently repeated three times. We report the mean accuracy and standard deviation as
\begin{equation}
\mu_{\text{Acc}} = \frac{1}{3} \sum_{r=1}^{3} \text{Acc}_r, \quad
\sigma_{\text{Acc}} = \sqrt{\frac{1}{3-1} \sum_{r=1}^{3} (\text{Acc}_r - \mu_{\text{Acc}})^2},
\end{equation}
where $\text{Acc}_r$ is the final cumulative accuracy of the $r$-th run.

\subsection{Parameter Study}

Before delving into specific algorithmic hyperparameters, we establish the default configurations used throughout the experiments. For initialization, the offline warm-up dataset contains exactly 1 sample per class to pre-train the base classifiers. The structural parameters of the base RVFL networks are set to $N_1 = 10$ and $N_2 = 10$. Furthermore, the ensemble size is fixed at $N = 10$, and the negative penalty parameter for complementary target construction is $\eta = -0.5$. Under these controlled settings, we investigate two critical hyperparameters governing the theoretical regret bound and adaptive tracking capability: the regret scaling factor $c \in \{0.05, 0.1, 0.5, 1.0, 5.0, 10.0\}$ and the periodic restart factor $\alpha \in [0.0, 1.0]$ with a step size of $0.1$. To ensure robustness, their interplay is evaluated under three feedback scarcity conditions: complete scarcity ($\rho = 0.0$), severe scarcity ($\rho = 0.1$), and moderate scarcity ($\rho = 0.5$). The resulting performance landscapes are visualized in Fig.~\ref{fig:parameter}.

Further analysis reveals that when $c$ is excessively large or $\alpha$ is too small, the exploration rate  $\gamma$ gets capped at 1. Consequently, the action sampling distribution under partial feedback degenerates into uniform random guessing, inducing severe performance degradation. Conversely, a smaller regret scaling factor ($c \in [0.05, 0.5]$) effectively prevents this $\gamma$ saturation, significantly enhancing predictive performance. Furthermore, the region of optimal accuracy consistently aligns with moderate-to-high restart factors ($\alpha \in [0.5, 0.9]$), indicating that periodic restarts effectively purge obsolete historical weights and facilitate rapid adaptation to concept drift. Intuitively, increasing the full-feedback probability $\rho$ supplies more ground-truth labels, consistently improving the overall accuracy. To achieve an optimal trade-off between stability and adaptability, we recommend $c \in [0.05, 0.5]$ and $\alpha \in [0.5, 0.9]$. Accordingly, we set $c=0.1$ and $\alpha=0.8$ in all subsequent comparative evaluations.

\begin{figure}[htbp]
    \centering
    \includegraphics[width=\linewidth]{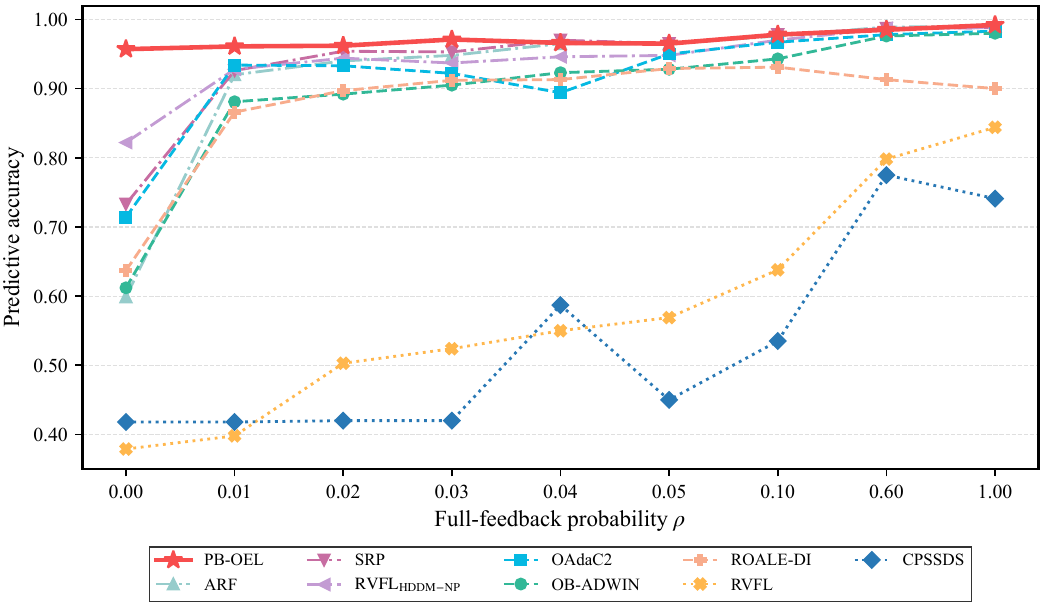}
    \caption{Predictive accuracy under different full-feedback probabilities $\rho$.}
    \label{fig:acc_rho}
\end{figure}

\begin{table*}[!t]
\begin{center}
  \centering
  \caption{Predictive accuracy (mean $\pm$ standard deviation) of the evaluated methods under varying full-feedback probabilities ($\bm{\rho}$) across 3 independent runs.}
  \label{tab:algorithm_comparison_rho}
  \begin{adjustbox}{width=\textwidth}
  \renewcommand{\arraystretch}{0.8} 
  \begin{tabular}{l *{9}{c}}
  \specialrule{0.15em}{1pt}{1pt}
  \specialrule{0.15em}{1pt}{3pt}

\multirow{2}{*}{\bf Ratio ($\bm{\rho}$)} & \multirow{2}{*}{\bf OB-ADWIN} & \multirow{2}{*}{\bf OAdaC2} & \multirow{2}{*}{\bf CPSSDS} & \multirow{2}{*}{\bf ARF}  & \multirow{2}{*}{\bf SRP} & \multirow{2}{*}{\bf ROALE-DI} & \multirow{2}{*}{\bf RVFL} & \multirow{2}{*}{$\textbf{RVFL}_{\textbf{HDDM-NP}}$} & \multirow{2}{*}{\bf PB-OEL*} \\
&&&&&&&&&\\
\cmidrule[0.5pt](r){1-10}

0.00  & 0.612 ± 0.045 & 0.714 ± 0.028 & 0.418 ± 0.000 & 0.599 ± 0.029 & 0.733 ± 0.032 & 0.637 ± 0.098 & 0.379 ± 0.015 & 0.822 ± 0.001 & \textbf{0.957 ± 0.000} \\
\cmidrule[0.5pt](r){1-10}

0.01  & 0.881 ± 0.002 & 0.934 ± 0.007 & 0.418 ± 0.000 & 0.920 ± 0.019 & 0.926 ± 0.015 & 0.866 ± 0.011 & 0.398 ± 0.038 & 0.928 ± 0.010 & \textbf{0.961 ± 0.000} \\
\cmidrule[0.5pt](r){1-10}

0.02  & 0.892 ± 0.001 & 0.933 ± 0.003 & 0.420 ± 0.000 & 0.940 ± 0.002 & 0.954 ± 0.006 & 0.897 ± 0.009 & 0.503 ± 0.018 & 0.944 ± 0.004 & \textbf{0.962 ± 0.000} \\
\cmidrule[0.5pt](r){1-10}

0.03  & 0.905 ± 0.005 & 0.922 ± 0.039 & 0.420 ± 0.000 & 0.948 ± 0.008 & 0.953 ± 0.004 & 0.912 ± 0.011 & 0.524 ± 0.024 & 0.937 ± 0.006 & \textbf{0.971 ± 0.000} \\
\cmidrule[0.5pt](r){1-10}

0.04  & 0.923 ± 0.001 & 0.894 ± 0.043 & 0.587 ± 0.072 & 0.965 ± 0.003 & \textbf{0.970 ± 0.002} & 0.913 ± 0.011 & 0.550 ± 0.011 & 0.946 ± 0.003 & 0.966 ± 0.000 \\
\cmidrule[0.5pt](r){1-10}

0.05  & 0.928 ± 0.002 & 0.950 ± 0.005 & 0.450 ± 0.020 & \textbf{0.965 ± 0.003} & \textbf{0.965 ± 0.009} & 0.929 ± 0.003 & 0.569 ± 0.008 & 0.948 ± 0.007 & \textbf{0.965 ± 0.000} \\
\cmidrule[0.5pt](r){1-10}

0.10  & 0.943 ± 0.007 & 0.967 ± 0.008 & 0.535 ± 0.081 & 0.976 ± 0.002 & \textbf{0.978 ± 0.005} & 0.931 ± 0.017 & 0.638 ± 0.035 & 0.970 ± 0.006 & \textbf{0.978 ± 0.000} \\
\cmidrule[0.5pt](r){1-10}

0.60  & 0.976 ± 0.001 & 0.978 ± 0.001 & 0.775 ± 0.025 & \textbf{0.989 ± 0.000} & 0.987 ± 0.000 & 0.913 ± 0.010 & 0.798 ± 0.005 & 0.987 ± 0.000 & 0.985 ± 0.000 \\
\cmidrule[0.5pt](r){1-10}

1.00  & 0.980 ± 0.001 & 0.983 ± 0.001 & 0.741 ± 0.000 & 0.990 ± 0.001 & 0.988 ± 0.000 & 0.900 ± 0.005 & 0.844 ± 0.017 & 0.988 ± 0.000 & \textbf{0.992 ± 0.000} \\
\cmidrule[0.5pt](r){1-10}
{\bf Avg. Acc} & 0.893 & 0.919 & 0.529 & 0.921 & 0.939 & 0.878 & 0.578 & 0.941 & \textbf{0.971} \\
\cmidrule[0.5pt](r){1-10}
{\bf Rank}     & 6 & 5 & 9 & 4 & 3 & 7 & 8 & 2 & \textbf{1} \\

  \specialrule{0.15em}{3pt}{1pt}
  \specialrule{0.15em}{1pt}{1pt}

\end{tabular}
\end{adjustbox}
\begin{tablenotes}\footnotesize
    \item[1] *Note: `*' represents the proposed approach. {The best performance} for each {full-feedback probability} is \textbf{highlighted in bold}.
\end{tablenotes}

 \end{center}
\end{table*}

\begin{figure*}[!t]
    \centering
    
    \begin{minipage}{0.33\textwidth}
        \centering
        \includegraphics[width=\linewidth]{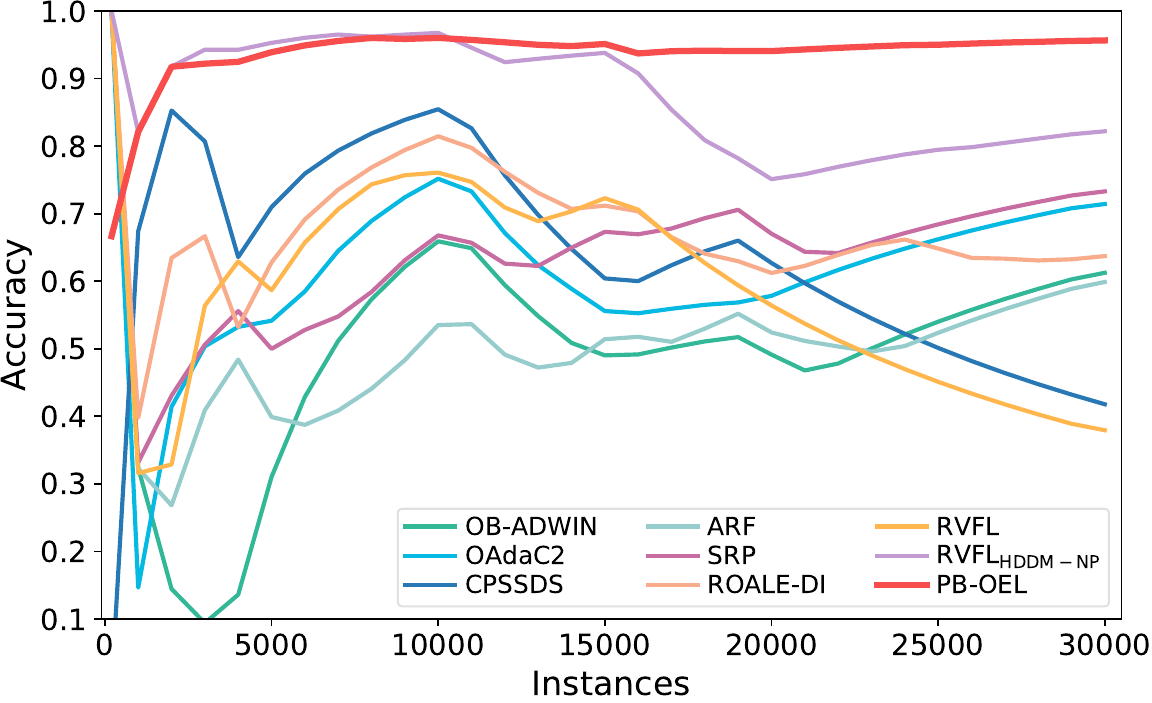}
        \centerline{(a) $\rho = 0.00$}
    \end{minipage}\hfill
    \begin{minipage}{0.33\textwidth}
        \centering
        \includegraphics[width=\linewidth]{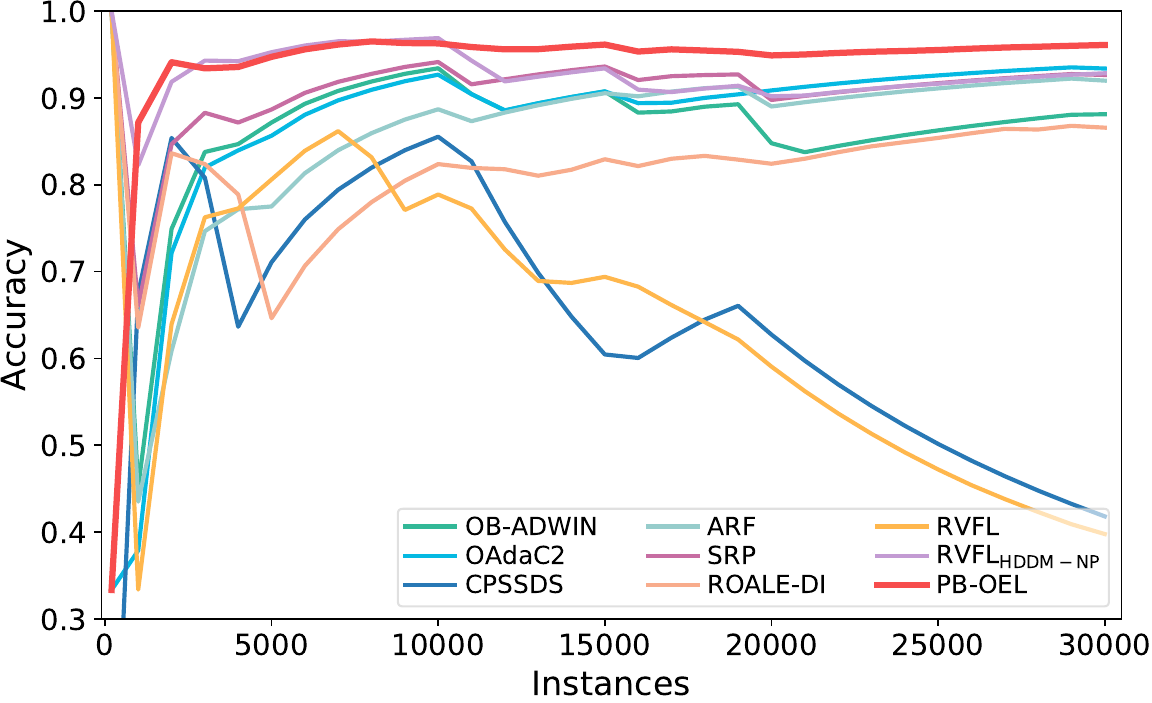}
        \centerline{(b) $\rho = 0.01$}
    \end{minipage}\hfill
    \begin{minipage}{0.33\textwidth}
        \centering
        \includegraphics[width=\linewidth]{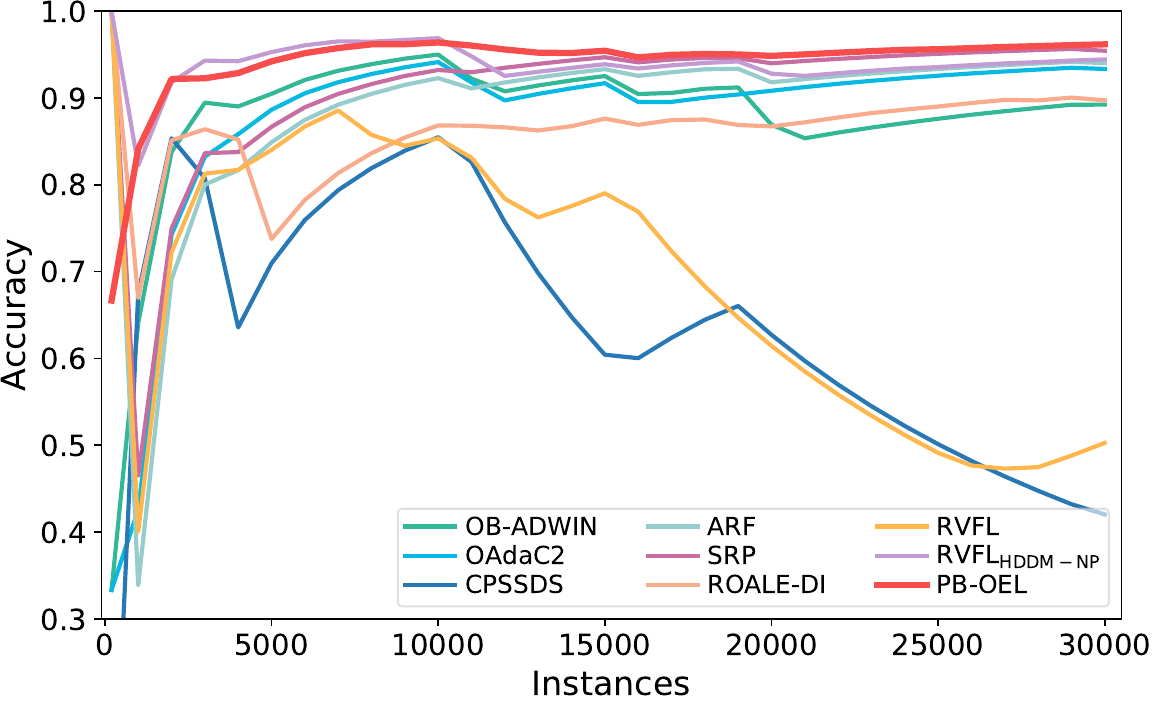}
        \centerline{(c) $\rho = 0.02$}
    \end{minipage}

    \begin{minipage}{0.33\textwidth}
        \centering
        \includegraphics[width=\linewidth]{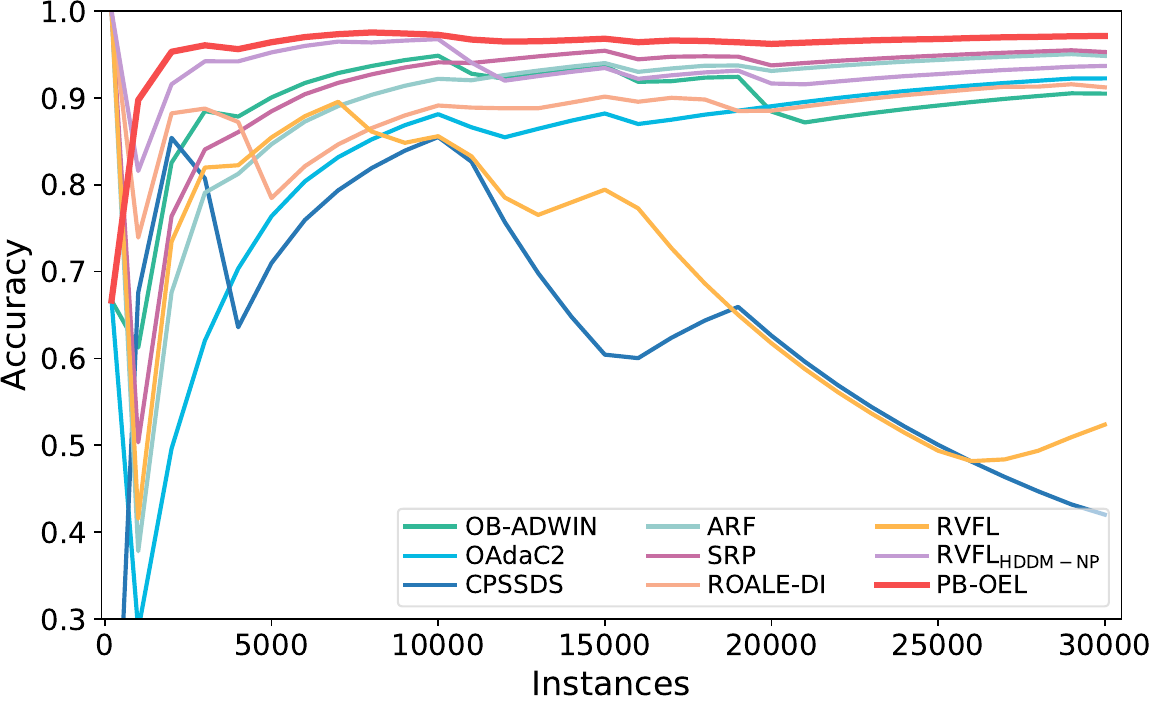}
        \centerline{(d) $\rho = 0.03$}
    \end{minipage}\hfill
    \begin{minipage}{0.33\textwidth}
        \centering
        \includegraphics[width=\linewidth]{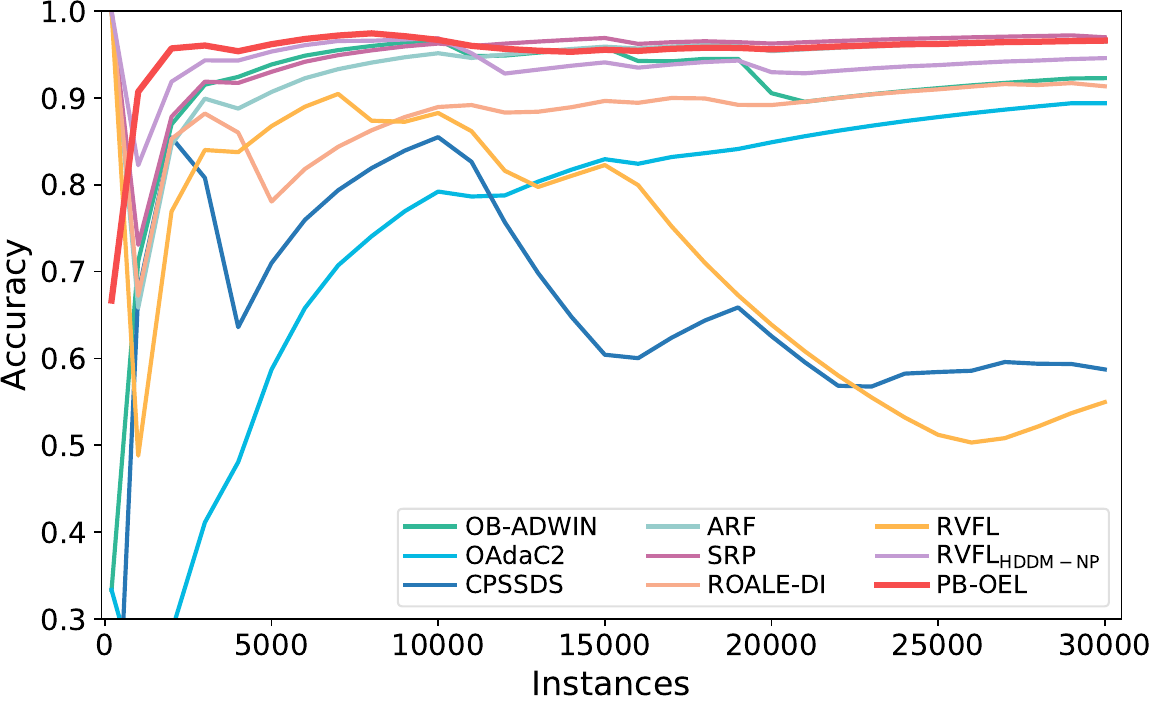}
        \centerline{(e) $\rho = 0.04$}
    \end{minipage}\hfill
    \begin{minipage}{0.33\textwidth}
        \centering
        \includegraphics[width=\linewidth]{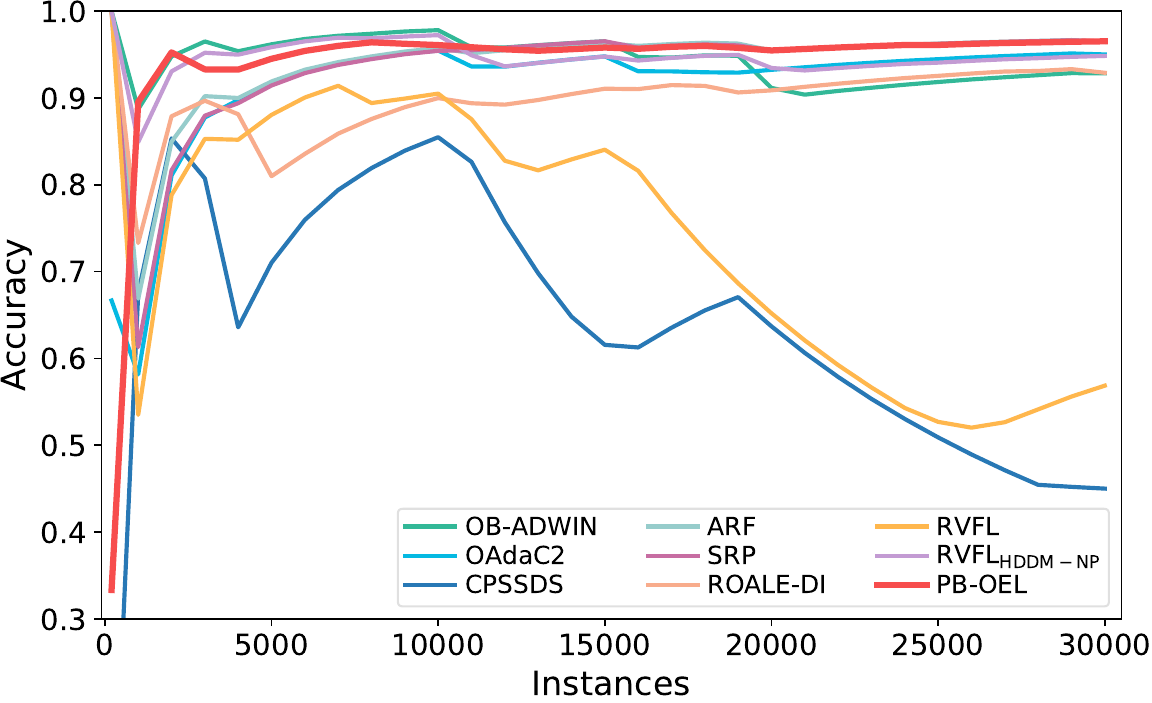}
        \centerline{(f) $\rho = 0.05$}
    \end{minipage}

    \begin{minipage}{0.33\textwidth}
        \centering
        \includegraphics[width=\linewidth]{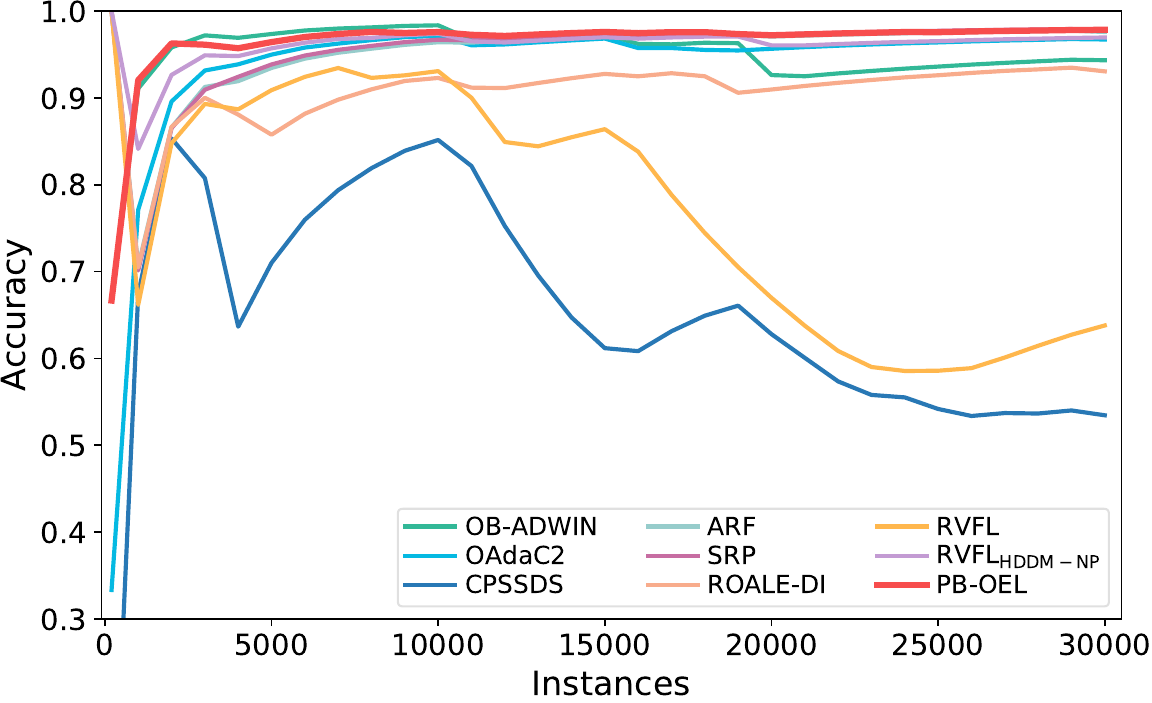}
        \centerline{(g) $\rho = 0.10$}
    \end{minipage}\hfill
    \begin{minipage}{0.33\textwidth}
        \centering
        \includegraphics[width=\linewidth]{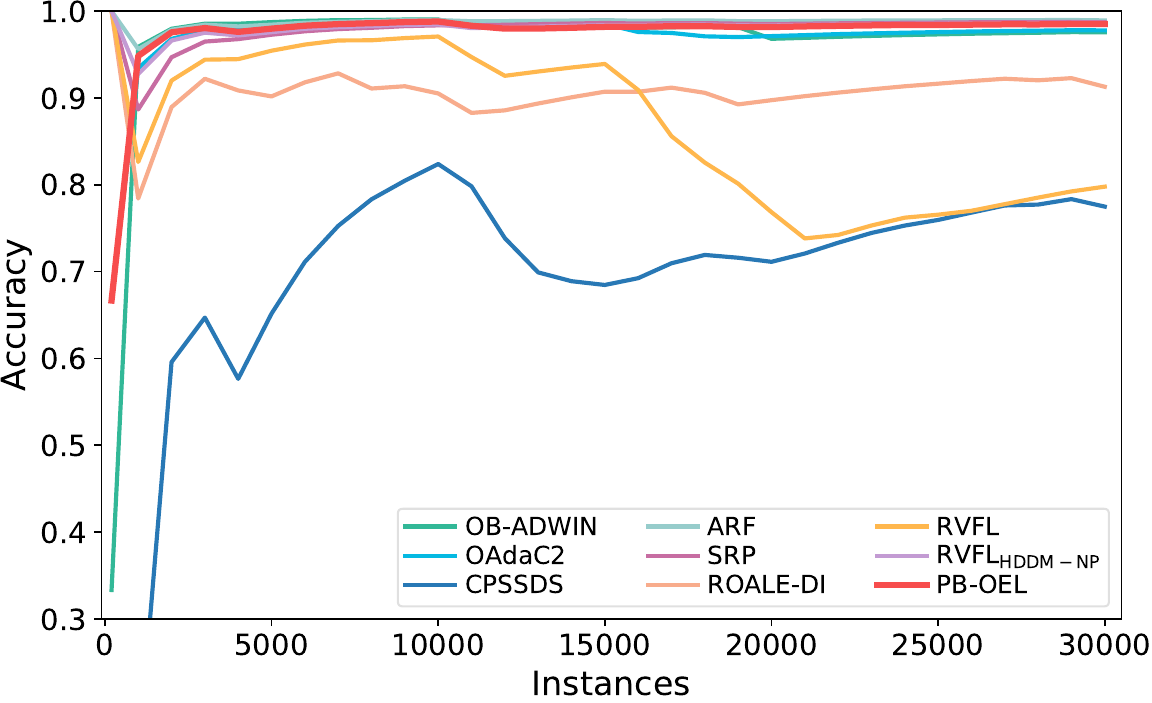}
        \centerline{(h) $\rho = 0.60$}
    \end{minipage}\hfill
    \begin{minipage}{0.33\textwidth}
        \centering
        \includegraphics[width=\linewidth]{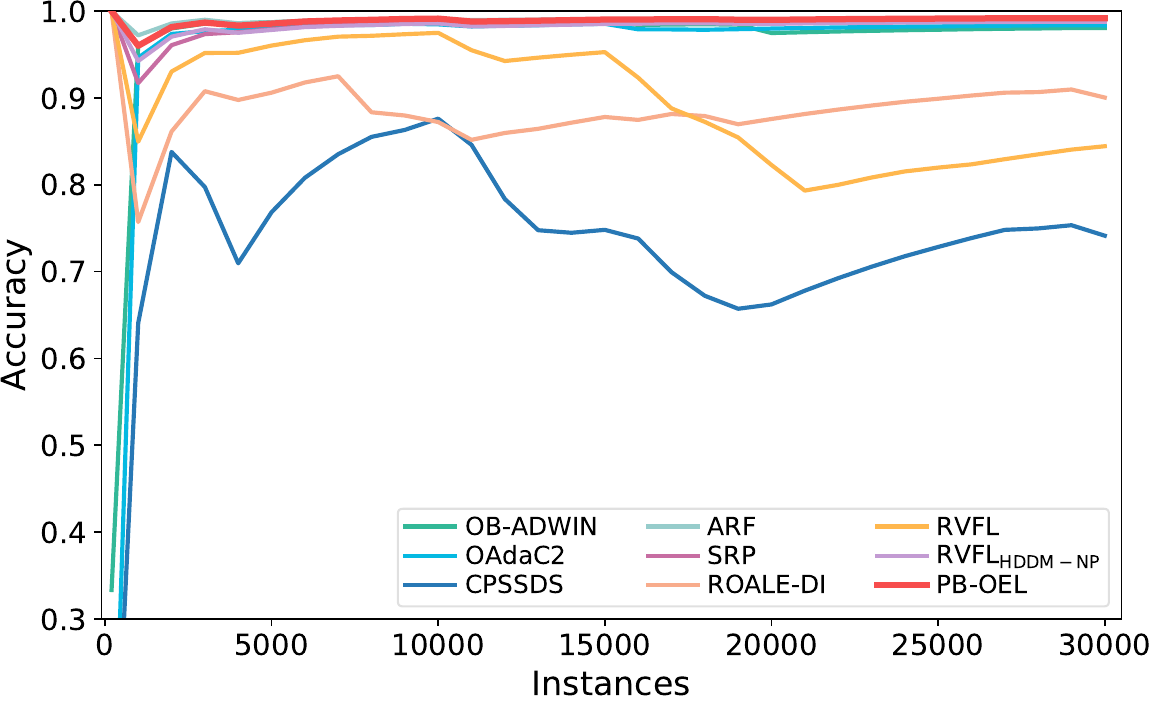}
        \centerline{(i) $\rho = 1.00$}
    \end{minipage}
    
    \caption{Prequential accuracy curves of the evaluated methods under varying full-feedback probabilities ($\bm{\rho}$) averaged over 3 independent runs.}
    \label{fig:all_rho_results}
\end{figure*}
\subsection{Comparative Evaluation}
To comprehensively evaluate the proposed PB-OEL framework under various mixed-feedback scenarios, we compare it against several state-of-the-art online learning algorithms\footnote{The source codes of the existing baseline methods are available at \url{https://github.com/liuzy0708/Awesome_OL}.} \cite{liu2025awesome}. The selected baselines encompass classical online ensemble methods such as OB-ADWIN \cite{oza2001online}, which integrates online bagging with the ADWIN drift detector, and OAdaC2 \cite{wang2016online}, a boosting-based approach incorporating unequal misclassification costs. We also evaluate robust tree-based and subspace ensembles, including Adaptive Random Forest (ARF) \cite{gomes2017adaptive} and Streaming Random Patches (SRP) \cite{gomes2019streaming}. Furthermore, we benchmark against recent advanced methods: ROALE-DI \cite{zhang2020reinforcement}, a framework that dynamically adjusts base classifier weights while handling data imbalance; and CPSSDS \cite{tanha2022cpssds}, which utilizes conformal prediction to enable drift detection with limited annotations. Notably, all aforementioned ensemble methods are implemented under their default parameter settings, with the ensemble size uniformly set to 10. Finally, for ablation purposes, we include two single-model baselines: a standard RVFL network, and $\text{RVFL}_{\text{HDDM-NP}}$, which equips the base RVFL solely with the HDDM drift detector and the negative penalty updating mechanism. Moreover, to prevent the models from degenerating into persistent suboptimal tracking under mixed-feedback conditions, an ${\epsilon}$-greedy exploration mechanism (${\epsilon = 0.005}$) is injected into the prediction phase of all evaluated methods. Detailed statistical results and averaged prequential accuracy curves are presented in Table~\ref{tab:algorithm_comparison_rho},  Fig.~\ref{fig:acc_rho} and Fig.~\ref{fig:all_rho_results}, respectively.

Baselines such as CPSSDS and the standard RVFL network exhibit inadequate drift adaptation capability under limited supervision, leading to severe performance degradation across most scenarios. Furthermore, classical online ensembles such as OB-ADWIN and OAdaC2, alongside dynamic weighting frameworks like ROALE-DI, suffer from significant instability in label-scarce environments; without sufficient ground-truth labels to correctly evaluate and update base classifiers, their prequential accuracy curves exhibit severe fluctuations. Meanwhile, tree-based ensembles like ARF and SRP perform exceptionally well under abundant feedback, achieving accuracies of 0.990 and 0.988, respectively, when the full-feedback probability reaches ${1.0}$. However, without sufficient ground-truth labels to supervise structural updates of the decision trees, they still struggle to recover promptly after concept drift occurs under severe label scarcity (${\rho \le 0.05}$). In contrast, the proposed PB-OEL effectively overcomes all these limitations. The ablation results confirm that while equipping the base RVFL with the HDDM detector and the negative penalty mechanism (denoted as $\text{RVFL}_{\text{HDDM-NP}}$) significantly boosts performance to the second-best rank, the complete PB-OEL framework further guarantees exceptional resilience. It matches the state-of-the-art ensembles at high feedback probabilities and maintains a robust accuracy of 0.957 even when completely devoid of full feedback (${\rho=0.0}$), successfully preventing the model from degenerating into persistent suboptimal tracking and ultimately achieving the highest overall average accuracy of 0.971.

\section{Conclusion}
\label{sec:conclusion}
Real-time safety assessment in complex dynamic systems is often hindered by the prohibitive cost of obtaining full safety labels, resulting in predominantly mixed-feedback scenarios. To address this critical challenge, this paper has proposed PB-OEL, a performance-bounded mixed-feedback online ensemble learning framework. Crucially, we have established a unified theoretical lower bound for the expected accuracy of the ensemble under these conditions. This bound mathematically guarantees that the ensemble asymptotically matches or exceeds the performance of the optimal hindsight base classifier as the length of the data stream approaches infinity. Extensive evaluations on the real-world \textit{Jiaolong} Deep-Sea Manned Submersible dataset have validated the effectiveness and demonstrated the superior robustness of PB-OEL against state-of-the-art methods under concept drift. Future work will explore tightening the theoretical bounds of the ensemble weighting mechanisms and extending the framework to accommodate scenarios such as active learning with limited annotations, noisy labels, and multi-label prediction.

\appendix
\subsection{Proof of Theorem~\ref{Theorem:ACC}}
\label{appendix:1}
Let $\eta=\gamma/K$, $W_t=\sum_{n=1}^{N}w_n(t)$, and $q_n(t)=w_n(t)/W_t$. The initial weights are $w_n(1)=1$, and hence $W_1=N$. The expert weights are updated by
\begin{equation}
\label{eq:proof_update}
w_n(t+1)=w_n(t)\exp\left(\eta\hat r_n(t)\right).
\end{equation}

For any fixed base classifier $n$, since $W_{T+1}\ge w_n(T+1)$, we have
\begin{equation}
\label{eq:proof_lower}
\ln\frac{W_{T+1}}{W_1}
\ge
\eta\sum_{t=1}^{T}\hat r_n(t)-\ln N.
\end{equation}
The reward estimator in Eq.~\eqref{eq:mu_hat_n} is unbiased under both feedback modes, i.e., $\mathbb{E}[\hat r_n(t)]=\xi_n^{y_t}(t)$.

Define the feedback subsequences as $\mathcal{T}_{\text{F}}=\{t:Z_t=\text{F}\}$ and $\mathcal{T}_{\text{P}}=\{t:Z_t=\text{P}\}$. On full-feedback rounds, $\hat r_n(t)=\xi_n^{y_t}(t)\in[0,1]$. Applying Hoeffding's lemma \cite{hoeffding1994probability} yields:
\begin{equation}
\label{eq:proof_F}
\begin{aligned}
\ln\frac{W_{t+1}}{W_t}
&=
\ln\sum_{n=1}^{N}q_n(t)
\exp\left(\eta\xi_n^{y_t}(t)\right)\\
&\le
\eta
\sum_{n=1}^{N}q_n(t)\xi_n^{y_t}(t)
+
\frac{\eta^2}{8}.
\end{aligned}
\end{equation}
Summing Eq.~\eqref{eq:proof_F} over $t\in\mathcal{T}_{\text{F}}$ gives
\begin{equation}
\label{eq:proof_F_sum}
\sum_{t\in\mathcal{T}_{\text{F}}}
\ln\frac{W_{t+1}}{W_t}
\le
\eta G_{\text{Std}}^{\text{F}}
+
\frac{\eta^2}{8}|\mathcal{T}_{\text{F}}|,
\end{equation}
where $G_{\text{Std}}^{\text{F}}=\sum_{t\in\mathcal{T}_{\text{F}}}\sum_{n=1}^{N}q_n(t)\xi_n^{y_t}(t)$. This term is the expected cumulative reward on full-feedback rounds, since the standard policy samples actions according to $s_k(t)=\sum_{n=1}^{N}q_n(t)\xi_n^k(t)$, and thus $\mathbb{E}_{a_t}[\mu_t(a_t) \mid Z_t = \text{F}]=s_{y_t}(t)=\sum_{n=1}^{N}q_n(t)\xi_n^{y_t}(t)$.

On partial-feedback rounds, the EXP4 importance-weighted estimator is used. Since $p_k(t)\ge\gamma/K$, we have $\eta\hat r_n(t)\le1$. Applying the standard reward-form EXP4 \cite{auer2002nonstochastic} analysis to the partial-feedback subsequence establishes the following bound for any fixed base classifier $n$:
\begin{equation}
\label{eq:proof_P}
\begin{aligned}
\sum_{t\in\mathcal{T}_{\text{P}}}
\xi_n^{y_t}(t)
-
\mathbb{E}_{a}
\left[
G_{\text{Std}}^{\text{P}}
\right]
&\le
\frac{K\ln N}{\gamma}\\
&\quad+
(e-1)\gamma
\sum_{t\in\mathcal{T}_{\text{P}}}
\xi_n^{y_t}(t),
\end{aligned}
\end{equation}
where $G_{\text{Std}}^{\text{P}}=\sum_{t\in\mathcal{T}_{\text{P}}}\mu_t(a_t)$ and $\mathbb{E}_{a}$ denotes expectation over action sampling.

Conditioned on the realization of feedback modes, the total expected cumulative reward of the standard policy satisfies
\begin{equation}
\mathbb{E}\!\left[{G_{\text{Std}}}(T)\right] = {G_{\text{Std}}^{\text{F}}} + \mathbb{E}_{a}\!\left[{G_{\text{Std}}^{\text{P}}} \right].
\end{equation}
Combining the full-feedback contribution in Eq.~\eqref{eq:proof_F_sum} and the partial-feedback EXP4 bound in Eq.~\eqref{eq:proof_P}, and leveraging the shared global potential to ensure the initial complexity penalty ($\ln N$) is incurred only once, it holds that for any fixed base classifier $n$,
\begin{equation}
\label{eq:proof_realized}
\begin{aligned}
\sum_{t=1}^{T}\xi_n^{y_t}(t)
-
\mathbb{E}
\left[
G_{\text{Std}}(T)
\right]
&\le
\frac{K\ln N}{\gamma}
+
\frac{\gamma}{8K}
|\mathcal{T}_{\text{F}}|\\
&\quad+
(e-1)\gamma
\sum_{t\in\mathcal{T}_{\text{P}}}
\xi_n^{y_t}(t).
\end{aligned}
\end{equation}

Since each sample provides full feedback with probability $\rho$ and partial feedback with probability $1-\rho$, and the feedback indicator is independent of the reward sequence, we have $\mathbb{E}[|\mathcal{T}_{\text{F}}|]=\rho T$ and $\mathbb{E}[\sum_{t\in\mathcal{T}_{\text{P}}}\xi_n^{y_t}(t)]=(1-\rho)\sum_{t=1}^{T}\xi_n^{y_t}(t)$. Taking the expectation over the independent Bernoulli feedback indicators in Eq.~\eqref{eq:proof_realized} yields
\begin{equation}
\label{eq:proof_regret}
\begin{aligned}
\sum_{t=1}^{T}\xi_n^{y_t}(t)
-
\mathbb{E}
\left[
G_{\text{Std}}(T)
\right]
&\le
\frac{K\ln N}{\gamma}
+
\frac{\rho\gamma T}{8K}\\
&\quad+
(1-\rho)(e-1)\gamma
\sum_{t=1}^{T}
\xi_n^{y_t}(t).
\end{aligned}
\end{equation}

Define the optimal hindsight expert as $n^\star=\arg\max_{1\le n\le N}\sum_{t=1}^{T}\xi_n^{y_t}(t)$ and its averaged accuracy as $A_T^\star=\frac{1}{T}\sum_{t=1}^{T}\xi_{n^\star}^{y_t}(t)$. Since $\mu_t(a_t)=\mathbf{1}\{a_t=y_t\}$, we have $\mathbb{E}[\text{ACC}_{\text{Std}}]=\mathbb{E}[G_{\text{Std}}(T)]/T$. Substituting $n=n^\star$ into Eq.~\eqref{eq:proof_regret}, dividing both sides by $T$, and rearranging gives
\begin{equation}
\label{eq:proof_acc_before_relax}
\begin{aligned}
\mathbb{E}
\left[
\text{ACC}_{\text{Std}}
\right]
&\ge
A_T^\star
-
\frac{K\ln N}{\gamma T}
-
\frac{\rho\gamma}{8K}\\
&\quad-
(1-\rho)(e-1)\gamma A_T^\star.
\end{aligned}
\end{equation}
Since $A_T^\star\le1$, we obtain
\begin{equation}
\label{eq:proof_acc}
\mathbb{E}
\left[
\text{ACC}_{\text{Std}}
\right]
\ge
A_T^\star
-
\frac{K\ln N}{\gamma T}
-
\gamma B_\rho,
\end{equation}
where $B_\rho=\rho/(8K)+(1-\rho)(e-1)$.

By choosing the optimal $\gamma=\sqrt{K\ln N/(TB_\rho)}<1$, the two penalty terms on the right-hand side of Eq.~\eqref{eq:proof_acc} are balanced, i.e., $K\ln N/(\gamma T)=\gamma B_\rho=\sqrt{B_\rho K\ln N/T}$. Therefore,
\[
\mathbb{E}
\left[
\text{ACC}_{\text{Std}}
\right]
\ge
A_T^\star
-
2\sqrt{
\frac{B_\rho K\ln N}{T}
}.
\]
Since $A_T^\star=\max_{1\le n\le N}\frac{1}{T}\sum_{t=1}^{T}\xi_n^{y_t}(t)$, Eq.~\eqref{eq:ACC} follows.

Finally, under a hard voting mechanism where $\xi_n^{y_t}(t) \in \{0, 1\}$, $A_T^\star$ exactly equates to the accuracy of the optimal base classifier in hindsight, denoted as $\text{ACC}_{n^\star}$. Because the theoretical penalty bounded by $2\sqrt{B_\rho K \ln N / T}$ strictly vanishes as $T \to \infty$, the expected accuracy of the ensemble asymptotically matches or exceeds that of the optimal expert, establishing the asymptotic limit in Eq.~(\ref{eq:ACC_limit}).
\hfill$\blacksquare$

\subsection{Proof of Corollary~\ref{corollary:restart}}
\label{appendix:2}
Because PB-OEL resets its expert weights at the beginning of each batch, the algorithm operates as an independent instance of the standard policy on every single batch $\mathcal{T}_j$. Applying Theorem \ref{Theorem:ACC} locally to a batch of length $\Delta_T$, the expected cumulative reward for this specific batch satisfies:
\begin{equation}
\mathbb{E}[G_{\text{PB-OEL}}(\mathcal{T}_j)] \ge \max_{1 \le n \le N} \sum_{t \in \mathcal{T}_j} \xi_n^{y_t}(t) - \text{BatchRegret}(\Delta_T).
\end{equation}

By substituting the scaled exploration rate $\gamma = c \gamma^\star$ into the expected regret bound derived in Theorem \ref{Theorem:ACC}, the theoretical penalty is amplified by a factor of $\frac{1}{2}(c + c^{-1})$. Thus, the cumulative regret for any individual batch is upper-bounded by $\frac{1}{2}(c + c^{-1}) \times 2\sqrt{B_\rho K \ln N \Delta_T} = (c + c^{-1})\sqrt{B_\rho K \ln N \Delta_T}$.

Summing the expected rewards and regrets over all $m_T = \lceil T/\Delta_T \rceil \le \frac{T}{\Delta_T} + 1$ batches, the global expected cumulative reward across the entire horizon $T$ satisfies:
\begin{equation}
\begin{aligned}
\mathbb{E}[G_{\text{PB-OEL}}(T)] &\ge \sum_{j=1}^{m_T} \max_{1\le n \le N} \sum_{t \in \mathcal{T}_j} \xi_{n}^{y_t}(t) \\
&\quad - m_T (c + c^{-1})\sqrt{B_\rho K \ln N \Delta_T}.
\end{aligned}
\end{equation}

Dividing this inequality by $T$ to obtain the averaged accuracy yields:
\begin{equation}
\mathbb{E}[\text{ACC}_{\text{PB-OEL}}] \ge \frac{1}{T}\sum_{j=1}^{m_T} \max_{1\le n \le N} \sum_{t \in \mathcal{T}_j} \xi_{n}^{y_t}(t) - {\mathcal{R}},
\end{equation}
where the regret penalty ${\mathcal{R}}$ is bounded by:
\begin{equation}
{\mathcal{R}} \le \frac{1}{T} \left( \frac{T}{\Delta_T} + 1 \right) (c + c^{-1})\sqrt{B_\rho K \ln N \Delta_T}.
\end{equation}

Distributing the terms and substituting the restart interval $\Delta_T = T^\alpha$, the penalty simplifies exactly to:
\begin{equation}
(c + c^{-1})\sqrt{B_\rho K \ln N} \left(T^{\frac{\alpha}{2}-1} + T^{-\frac{\alpha}{2}}\right).
\end{equation}

Since $c$ is a constant independent of $T$, and for any $\alpha \in (0, 1]$, both exponent terms ($\frac{\alpha}{2}-1$ and $-\frac{\alpha}{2}$) are strictly negative. Consequently, the scaled piecewise regret preserves its asymptotic convergence to zero, completing the proof. $\hfill\blacksquare$

\subsection{Proof of Corollary~\ref{coro:comparison}}
\label{appendix:3}
{\itshape Part 1 (Higher Bound)}. Let $B_1$ and $B_2$ denote the {\itshape Piecewise Ultimate Bound} of PB-OEL and the {\itshape Standard Ultimate Bound} of the scaled standard policy, respectively. Similar to the proof of Corollary \ref{corollary:restart}, partition the time sequence $\mathcal{T}$ into $m_T$ distinct batches $\mathcal{T}_1, \cdots, \mathcal{T}_{m_T}$. Define:
\begin{equation*}
n_0 = \mathop{\operatorname{argmax}}\limits_{n \in \{1, 2, \cdots, N\}} \sum_{t \in \mathcal{T}} \xi_{n}^{y_t}(t)
\end{equation*}
which represents the hindsight global optimal expert with the best expected performance over the entire horizon $\mathcal{T}$. Then $B_1$ and $B_2$ can be compactly expressed as:
\begin{equation}
\left\{
\begin{array}{l}
\begin{aligned}
B_1 &= \frac{1}{T}\left(\max\limits_{n} \sum\limits_{t \in \mathcal{T}_1} \xi_{n}^{y_t}(t) + \cdots + \max\limits_{n} \sum\limits_{t \in \mathcal{T}_{m_T}} \xi_{n}^{y_t}(t) \right), \\
B_2 &= \frac{1}{T}\left({\sum\limits_{t \in \mathcal{T}_1} \xi_{n_0}^{y_t}(t)+\cdots+\sum\limits_{t \in \mathcal{T}_{m_T}} \xi_{n_0}^{y_t}(t)}\right).
\end{aligned}
\end{array}
\right.
\end{equation}

By the definition of the maximum operator, the maximum over any given subset is bounded below by the value of any specific element within that subset. Thus:
\begin{equation}
\max\limits_{1 \leq n \leq N}\sum_{t \in \mathcal{T}_j} \xi_{n}^{y_t}(t) \geq \sum_{t \in \mathcal{T}_j} \xi_{n_0}^{y_t}(t), \quad \forall \mathcal{T}_j \in \{\mathcal{T}_1, \mathcal{T}_2, \cdots, \mathcal{T}_{m_T}\},
\end{equation}
which directly leads to $B_1 \geq B_2$ upon summation over all $m_T$ batches.

{\itshape Part 2 (Greater Regret)}. Let $R_1$ and $R_2$ represent the {\itshape Scaled Piecewise Regret} of PB-OEL and the {\itshape Scaled Standard Regret} of the scaled standard policy, respectively:
\begin{equation}
\begin{aligned}
R_1 &= (c+c^{-1}) \sqrt{B_\rho K \ln N} \left(T^{\frac{\alpha}{2}-1} + T^{-\frac{\alpha}{2}}\right), \\
R_2 &= (c+c^{-1}) \sqrt{B_\rho K \ln N} \cdot T^{-\frac{1}{2}}.
\end{aligned}
\end{equation}

Subtracting $R_2$ from $R_1$ and factoring out $T^{-\frac{1}{2}}$ yields:
\begin{equation}
\label{eq:difference}
\begin{aligned}
R_1 - R_2 &= \frac{(c+c^{-1}) \sqrt{B_\rho K \ln N}}{\sqrt{T}}\left(T^{\frac{1-\alpha}{2}} + T^{-\frac{1-\alpha}{2}} - 1\right).
\end{aligned}
\end{equation}

By the Arithmetic Mean-Geometric Mean (AM-GM) inequality, it trivially holds that for any $T \ge 1$:
\begin{equation}
\label{eq:difference2}
\begin{aligned}
T^{\frac{1-\alpha}{2}} + T^{-\frac{1-\alpha}{2}} \geq 2, \quad \forall \alpha \in (0, 1],
\end{aligned}
\end{equation}
which ensures $R_1 - R_2 \ge {\frac{(c+c^{-1}) \sqrt{B_\rho K \ln N}}{\sqrt{T}}} > 0$, hence $R_1 > R_2$. Note that even in the extreme case where $\alpha=1$ (i.e., no actual restart during $T$), $R_1$ remains strictly greater than $R_2$. This minor discrepancy fundamentally arises from the ceiling relaxation $m_T = \lceil T/{\Delta_T} \rceil \le T/{\Delta_T} + 1$ utilized in Corollary \ref{corollary:restart}, which introduces an additive unit term to strictly upper-bound the potentially incomplete final batch.$\hfill\blacksquare$

\bibliographystyle{ieeetr}
\bibliography{References}

@article{besbes2014stochastic,
  title={Stochastic multi-armed-bandit problem with non-stationary rewards},
  author={Besbes, Omar and Gur, Yonatan and Zeevi, Assaf},
  journal={Advances in neural information processing systems},
  volume={27},
  year={2014}
}

@article{auer2002nonstochastic,
  title={The nonstochastic multiarmed bandit problem},
  author={Auer, Peter and Cesa-Bianchi, Nicolo and Freund, Yoav and Schapire, Robert E},
  journal={SIAM journal on computing},
  volume={32},
  number={1},
  pages={48--77},
  year={2002},
  publisher={SIAM}
}

@article{frias2014online,
  title={Online and non-parametric drift detection methods based on {{Hoeffding}}’s bounds},
  author={Frias-Blanco, Isvani and others},
  journal={IEEE Transactions on Knowledge and Data Engineering},
  volume={27},
  number={3},
  pages={810--823},
  year={2014},
  publisher={IEEE}
}

@article{hoeffding1994probability,
  title={Probability inequalities for sums of bounded random variables},
  author={Hoeffding, Wassily},
  journal={The Collected Works of Wassily Hoeffding},
  pages={409--426},
  year={1994},
  publisher={Springer}
}

@book{lattimore2020bandit,
  title={Bandit algorithms},
  author={Lattimore, Tor and Szepesv{\'a}ri, Csaba},
  year={2020},
  publisher={Cambridge University Press}
}

@article{lu2019adaptive,
  title={Adaptive chunk-based dynamic weighted majority for imbalanced data streams with concept drift},
  author={Lu, Yang and Cheung, Yiuming and Tang, Yuanyan},
  journal={IEEE Transactions on Neural Networks and Learning Systems},
  volume={31},
  number={8},
  pages={2764--2778},
  year={2019},
  publisher={IEEE}
}

@article{gomes2017adaptive,
  title={Adaptive random forests for evolving data stream classification},
  author={Gomes, Heitor M and others},
  journal={Machine Learning},
  volume={106},
  pages={1469--1495},
  year={2017},
  publisher={Springer}
}

@article{zhang2020reinforcement,
  title={Reinforcement online active learning ensemble for drifting imbalanced data streams},
  author={Zhang, Hang and Liu, Weike and Liu, Qingbao},
  journal={IEEE Transactions on Knowledge and Data Engineering},
  volume={34},
  number={8},
  pages={3971--3983},
  year={2020},
  publisher={IEEE}
}

@inproceedings{gomes2019streaming,
  title={Streaming random patches for evolving data stream classification},
  author={Gomes, Heitor Murilo and Read, Jesse and Bifet, Albert},
  booktitle={2019 IEEE International Conference on Data Mining (ICDM)},
  pages={240--249},
  year={2019},
  organization={Beijing, China, IEEE}
}

@article{pao1994learning,
  title={Learning and generalization characteristics of the random vector functional-link net},
  author={Pao, Yoh-Han and Park, Gwang-Hoon and Sobajic, Dejan J},
  journal={Neurocomputing},
  volume={6},
  number={2},
  pages={163--180},
  year={1994},
  publisher={Elsevier}
}

@article{zhang2019online,
  title={Online learning method for drift and imbalance problem in client credit assessment},
  author={Zhang, Hang and Liu, Qingbao},
  journal={Symmetry},
  volume={11},
  number={7},
  pages={890},
  year={2019},
  publisher={MDPI}
}

@article{vzliobaite2016overview,
  title={An overview of concept drift applications},
  author={{\v{Z}}liobait{\.e}, Indr{\.e} and Pechenizkiy, Mykola and Gama, Joao},
  journal={Big Data Analysis: New Algorithms for A New Society},
  pages={91--114},
  year={2016},
  publisher={Springer}
}

@article{krawczyk2017ensemble,
  title={Ensemble learning for data stream analysis: A survey},
  author={Krawczyk, Bartosz and Minku, Leandro L and Gama, Jo{\~a}o and Stefanowski, Jerzy and Wo{\'z}niak, Micha{\l}},
  journal={Information Fusion},
  volume={37},
  pages={132--156},
  year={2017},
  publisher={Elsevier}
}

@article{cano2020kappa,
  title={Kappa updated ensemble for drifting data stream mining},
  author={Cano, Alberto and Krawczyk, Bartosz},
  journal={Machine Learning},
  volume={109},
  number={1},
  pages={175--218},
  year={2020},
  publisher={Springer}
}

@article{ren2018knowledge,
  title={Knowledge-maximized ensemble algorithm for different types of concept drift},
  author={Ren, Siqi and Liao, Bo and Zhu, Wen and Li, Keqin},
  journal={Information Sciences},
  volume={430},
  pages={261--281},
  year={2018},
  publisher={Elsevier}
}

@inproceedings{pang2018dynamic,
  title={Dynamic ensemble active learning: A non-stationary bandit with expert advice},
  author={Pang, Kunkun and Dong, Mingzhi and Wu, Yang and Hospedales, Timothy M},
  booktitle={2018 24th ICPR},
  pages={2269--2276},
  year={2018},
  organization={Beijing, China, IEEE}
}

@article{wilsonmulti,
  title={Multi-armed bandit based online model selection for concept-drift adaptation},
  author={Wilson, Jobin and Chaudhury, Santanu and Lall, Brejesh},
  journal={Expert Systems},
  year={2024},
  pages={e13626},
  publisher={Wiley Online Library}
}

@article{tekin2016adaptive,
  title={Adaptive ensemble learning with confidence bounds},
  author={Tekin, Cem and Yoon, Jinsung and Van Der Schaar, Mihaela},
  journal={IEEE Transactions on Signal Processing},
  volume={65},
  number={4},
  pages={888--903},
  year={2016},
  publisher={IEEE}
}

@article{lu2018learning,
  title={Learning under concept drift: A review},
  author={Lu, Jie and others},
  journal={IEEE Transactions on Knowledge and Data Engineering},
  volume={31},
  number={12},
  pages={2346--2363},
  year={2018},
  publisher={IEEE}
}

@inproceedings{oza2001online,
  title={Online bagging and boosting},
  author={Oza, Nikunj C and Russell, Stuart J},
  booktitle={International Workshop on Artificial Intelligence and Statistics},
  pages={229--236},
  year={2001},
  organization={Florida, USA, PMLR}
}

@article{wang2016online,
  title={Online bagging and boosting for imbalanced data streams},
  author={Wang, Boyu and Pineau, Joelle},
  journal={IEEE Transactions on Knowledge and Data Engineering},
  volume={28},
  number={12},
  pages={3353--3366},
  year={2016},
  publisher={IEEE}
}

@article{wilson2023homogeneous,
  title={Homogeneous--Heterogeneous Hybrid Ensemble for concept-drift adaptation},
  author={Wilson, Jobin and Chaudhury, Santanu and Lall, Brejesh},
  journal={Neurocomputing},
  volume={557},
  pages={126741},
  year={2023},
  publisher={Elsevier}
}

@article{liu2023robust,
  title={Robust Sequential Online Prediction with Dynamic Ensemble of Multiple Models: A Review},
  author={Liu, Bin},
  journal={Neurocomputing},
  pages={126553},
  year={2023},
  publisher={Elsevier}
}

@article{tanveer2023ensemble,
  title={Ensemble deep learning in speech signal tasks: A review},
  author={Tanveer, M and others},
  journal={Neurocomputing},
  pages={126436},
  year={2023},
  publisher={Elsevier}
}

@article{li2022high,
  title={High-dimensional multi-label data stream classification with concept drifting detection},
  author={Li, Peipei and Zhang, Haixiang and Hu, Xuegang and Wu, Xindong},
  journal={IEEE Transactions on Knowledge and Data Engineering},
  year={2023},
  volume={35},
  number={8},
  pages={8085-8099},
  publisher={IEEE}
}

@article{liu2022concept,
  title={Concept drift detection delay index},
  author={Liu, Anjin and Lu, Jie and Song, Yiliao and Xuan, Junyu and Zhang, Guangquan},
  journal={IEEE Transactions on Knowledge and Data Engineering},
  volume={35},
  number={5},
  pages={4585--4597},
  year={2022},
  publisher={IEEE}
}

@article{liu2022online,
  author={Liu, Zeyi and Zhang, Yi and Ding, Zhongjun and He, Xiao},
  journal={IEEE Transactions on Neural Networks and Learning Systems}, 
  title={An Online Active Broad Learning Approach for Real-Time Safety Assessment of Dynamic Systems in Nonstationary Environments}, 
  year={2023},
  volume={34},
  number={10},
  pages={6714-6724}}

@article{hu2024cadm,
  title={{CADM$+$}: Confusion-Based Learning Framework With Drift Detection and Adaptation for Real-Time Safety Assessment},
  author={Hu, Songqiao and Liu, Zeyi and Li, Minyue and He, Xiao},
  journal={IEEE Transactions on Neural Networks and Learning Systems},
  year={2025},
  volume={36},
  number={3},
  pages={5126-5139},
  publisher={IEEE}
}

@article{liu2023dynamic,
  title={Dynamic submodular-based learning strategy in imbalanced drifting streams for real-time safety assessment in nonstationary environments},
  author={Liu, Zeyi and He, Xiao},
  journal={IEEE Transactions on Neural Networks and Learning Systems},
  year={2024},
  volume={35},
  number={3},
  pages={3038-3051},
  publisher={IEEE}
}

@article{he2023dynamic,
  author={He, Xiao and Liu, Zeyi},
  journal={IEEE Transactions on Cybernetics}, 
  title={Dynamic Model Interpretation-Guided Online Active Learning Scheme for Real-Time Safety Assessment}, 
  year={2024},
  volume={54},
  number={5},
  pages={2734-2745},
  }

@article{campagner2024ensemble,
  title={Ensemble predictors: Possibilistic combination of conformal predictors for multivariate time series classification},
  author={Campagner, Andrea and Barandas, Mar{\'\i}lia and Folgado, Duarte and Gamboa, Hugo and Cabitza, Federico},
  journal={IEEE Transactions on Pattern Analysis and Machine Intelligence},
   year={2024},
  volume={46},
  number={11},
  pages={7205-7216},
  publisher={IEEE}
}

@article{zhou2024class,
  title={Class-incremental learning: A survey},
  author={Zhou, Dawei and Wang, Qiwei and Qi, Zhihong and Ye, Hanjia and Zhan, Dechuan and Liu, Ziwei},
  journal={IEEE Transactions on Pattern Analysis and Machine Intelligence},
  year={2024},
  volume={46},
  number={12},
  pages={9851-9873},
  publisher={IEEE}
}

@article{tanha2022cpssds,
  title={CPSSDS: Conformal prediction for semi-supervised classification on data streams},
  author={Tanha, Jafar and Samadi, Negin and Abdi, Yousef and Razzaghi-Asl, Nazila},
  journal={Information Sciences},
  volume={584},
  pages={212--234},
  year={2022},
  publisher={Elsevier}
}

@article{he2024real,
  title={Real-time safety assessment techniques of dynamic systems},
  author={X. He and Z. Liu and S. Hu and C. Liu and D. Zhou},
  journal={Acta Automatica Sinica},
  volume={51},
  number={2},
  pages={249--270},
  year={2024},
  publisher={Beijing Zhongke Journal Publising Co. Ltd.}
}

@article{cacciarelli2024active,
  title={Active learning for data streams: a survey},
  author={Cacciarelli, Davide and Kulahci, Murat},
  journal={Machine Learning},
  volume={113},
  number={1},
  pages={185--239},
  year={2024},
  publisher={Springer}
}

@article{lin2025uncertainty,
  title={Uncertainty-Driven Online Deep Ensemble for Imbalanced Drifting Data Stream Regression},
  author={Lin, Yanhui and Qi, Lin},
  journal={IEEE Transactions on Industrial Informatics},
  year={2026},
  volume={22},
  number={1},
  pages={49-59},
  publisher={IEEE}
}

@article{freund1997decision,
  title={A decision-theoretic generalization of on-line learning and an application to boosting},
  author={Freund, Yoav and Schapire, Robert E},
  journal={Journal of computer and system sciences},
  volume={55},
  number={1},
  pages={119--139},
  year={1997},
  publisher={Elsevier}
}

@article{zhou2018brief,
  title={A brief introduction to weakly supervised learning},
  author={Zhou, Zhihua},
  journal={National science review},
  volume={5},
  number={1},
  pages={44--53},
  year={2018},
  publisher={Oxford University Press}
}

@article{ishida2017learning,
  title={Learning from complementary labels},
  author={Ishida, Takashi and Niu, Gang and Hu, Weihua and Sugiyama, Masashi},
  journal={Advances in neural information processing systems},
  volume={30},
  year={2017}
}

@article{hager1989updating,
  title={Updating the inverse of a matrix},
  author={Hager, William W},
  journal={SIAM review},
  volume={31},
  number={2},
  pages={221--239},
  year={1989},
  publisher={SIAM}
}

@article{liu2025awesome,
  title={Awesome-OL: An Extensible Toolkit for Online Learning},
  author={Liu, Zeyi and Hu, Songqiao and Han, Pengyu and Liu, Jiaming and He, Xiao},
  journal={arXiv preprint arXiv:2507.20144},
  year={2025}
}

@article{jiao2022dynamic,
  title={Dynamic ensemble selection for imbalanced data streams with concept drift},
  author={Jiao, Botao and Guo, Yinan and Gong, Dunwei and Chen, Qiuju},
  journal={IEEE Transactions on Neural Networks and Learning Systems},
  volume={35},
  number={1},
  pages={1278--1291},
  year={2022},
  publisher={IEEE}
}

@article{yan2024incremental,
  title={Incremental model evolution for power system security early warning based on knowledge distillation and active learning},
  author={Yan, Jiongcheng and Li, Changgang and Liu, Yutian and Yu, Dongxiao and Jia, Zhiping},
  journal={IEEE Transactions on Industrial Informatics},
  volume={20},
  number={11},
  pages={12958--12968},
  year={2024},
  publisher={IEEE}
}

@article{cao2023online,
  author={Cao, Chaojin and He, Yaoyao},
  journal={IEEE Transactions on Industrial Informatics}, 
  title={An Online Probability Density Load Forecasting Against Concept Drift Under Anomalous Events}, 
  year={2024},
  volume={20},
  number={4},
  pages={5241-5252}}

@article{qian2024deep,
  title={Deep imbalanced separation network: a holistic fault detection framework considering class-imbalance and partial label-unknown},
  author={Qian, Min and Li, Yanfu and Wu, Hui},
  journal={IEEE Transactions on Industrial Informatics},
  volume={20},
  number={11},
  pages={13026--13035},
  year={2024},
  publisher={IEEE}
}

@article{sun2024novel,
  author={Sun, Meidi and Xiao, Xinmiao and Chen, Tangyan and He, Qing and Long, Zhuo and Li, Ling},
  journal={IEEE Transactions on Industrial Informatics}, 
  title={A Novel Domain Incremental Learning Method for Bearing Fault Diagnosis Based on {F\&K}}, 
  year={2025},
  volume={21},
  number={1},
  pages={980-989}}

@article{yan2024multisensor,
  author={Yan, Xunshi and Shi, Zhengang and Sun, Zhe and Zhang, Chen-An},
  journal={IEEE Transactions on Industrial Informatics}, 
  title={Multisensor Fusion on Hypergraph for Fault Diagnosis}, 
  year={2024},
  volume={20},
  number={8},
  pages={10008-10018}}
\end{document}